\documentclass{article}

\usepackage{PRIMEarxiv}

\usepackage[utf8]{inputenc} %
\usepackage[T1]{fontenc}    %
\usepackage{hyperref}       %
\usepackage{url}            %
\usepackage{booktabs}       %
\usepackage{amsfonts}       %
\usepackage{nicefrac}       %
\usepackage{microtype}      %
\usepackage{lipsum}
\usepackage{fancyhdr}       %
\usepackage{graphicx}       %
\graphicspath{{media/}}     %

\usepackage{natbib}
\usepackage[textsize=tiny]{todonotes}
\usepackage{hyperref}

\usepackage{amsmath}
\usepackage{amssymb}
\usepackage{mathtools}
\usepackage{amsthm}

\usepackage{multirow}

\hypersetup{
colorlinks,
linkcolor=blue
,citecolor=black
,filecolor=black
,urlcolor=blue
,menucolor=black
,runcolor=black
}

\usepackage[capitalize, nameinlink, noabbrev]{cleveref}

\usepackage{subcaption} %
\RequirePackage{algorithm}
\RequirePackage{algorithmic}

\usepackage{xspace}

\hypersetup{
colorlinks,
linkcolor=NavyBlue
,citecolor=PineGreen
,filecolor=Mulberry
,urlcolor=NavyBlue
,menucolor=BrickRed
,runcolor=Mulberry
}

\title{Stein Variational Newton Neural Network Ensembles
}

\usepackage{etoolbox}
\makeatletter
\patchcmd{\maketitle}
 {\def\@makefnmark}
 {\def\@makefnmark{}\def\useless@macro}
 {}{}
\makeatother

\newif\ifuniqueAffiliation

\ifuniqueAffiliation %
\author{
	{Klemens Flöge} \\
    Helmholtz AI, Munich \\
	\texttt{klemens.floege@gmail.com} \\
    \And
    {Mohammed Abdul Moeed} \\
	Department of Computer Science\\
	Technical University of Munich\\
    \And
    {Vincent Fortuin} \\
	Department of Computer Science\\
	Technical University of Munich\\
	Helmholtz AI, Munich \\
	\texttt{vincent.fortuin@helmholtz-munich.de} \\
}
\else
\usepackage{authblk}

\author[1]{%
    {Klemens Flöge\thanks{Correspondence to \texttt{klemens.floege@gmail.com}}\hspace{0.4em}}%
}
\author[2]{%
	{Mohammed Abdul Moeed}%
}
\author[1,2,3]{%
	{Vincent Fortuin}%
}
\affil[1]{Helmholtz AI, Munich}
\affil[2]{Department of Computer Science, Technical University of Munich}
\affil[3]{Munich Center for Machine Learning}
\fi

\begin{document}
\maketitle

\begin{abstract}
Deep neural network ensembles are powerful tools for uncertainty quantification, which have recently been re-interpreted from a Bayesian perspective.
However, current methods inadequately leverage second-order information of the loss landscape, despite the recent availability of efficient Hessian approximations.
We propose a novel approximate Bayesian inference method that modifies deep ensembles to incorporate Stein Variational Newton updates.
Our approach uniquely integrates scalable modern Hessian approximations, achieving faster convergence and more accurate posterior distribution approximations.
We validate the effectiveness of our method on diverse regression and classification tasks, demonstrating superior performance with a significantly reduced number of training epochs compared to existing ensemble-based methods, while enhancing uncertainty quantification and robustness against overfitting.\footnote[2]{Our code is available here: \href{https://github.com/klemens-floege/svn_ensembles}{https://github.com/klemens-floege/svn\_ensembles}}

\end{abstract}


\section{Introduction}
\label{sec:introduction}

Effectively approximating intractable probability distributions with finite samples is crucial for evaluating expectations and characterizing uncertainties in machine learning and statistics \citep{Pearson1895, Silverman86, givens2012computational}. This challenge is particularly pronounced in high-dimensional and multimodal distributions, such as those in approximate Bayesian inference for deep neural networks. Traditional methods to address these challenges include parametric variational inference \citep[VI;][]{blei2017variational} and Markov chain Monte Carlo \citep[MCMC;][]{andrieu2003introduction}, each with tradeoffs between computational efficiency and estimation accuracy.

Stein Variational Gradient Descent \citep[SVGD;][]{liu2016stein} is a potent non-parametric VI technique that balances efficiency and performance by coupling a tractable reference distribution with a complex target distribution via transport maps. Enhancing SVGD with curvature information results in Stein Variational Newton \citep[SVN;][]{detommaso2018stein}, which, despite the increased cost of Hessian computations, achieves significantly faster convergence. However, until now, SVN's application has been limited to low-dimensional inference problems.

Since the advent of neural networks, researchers have applied Bayesian principles to them \citep[BNNs;][]{neal1995bayesian}. SVGD has also been extended to neural networks \citep{d2021stein}, however, for SVN, this still remains elusive.
\citet{li2018visualizing} showed that deep neural network loss landscapes are complex yet structured \citep{garipov2018loss}, suggesting that curvature information can improve learning dynamics \citep{lin2023structured, lin2024remove, eschenhagen2024kroneckerfactored}. Therefore, we aim to extend SVN to high-dimensional neural network posteriors.

\begin{figure}
    \centering
    \begin{minipage}[b]{0.35\textwidth}
        \centering
        \includegraphics[width=0.99\textwidth]{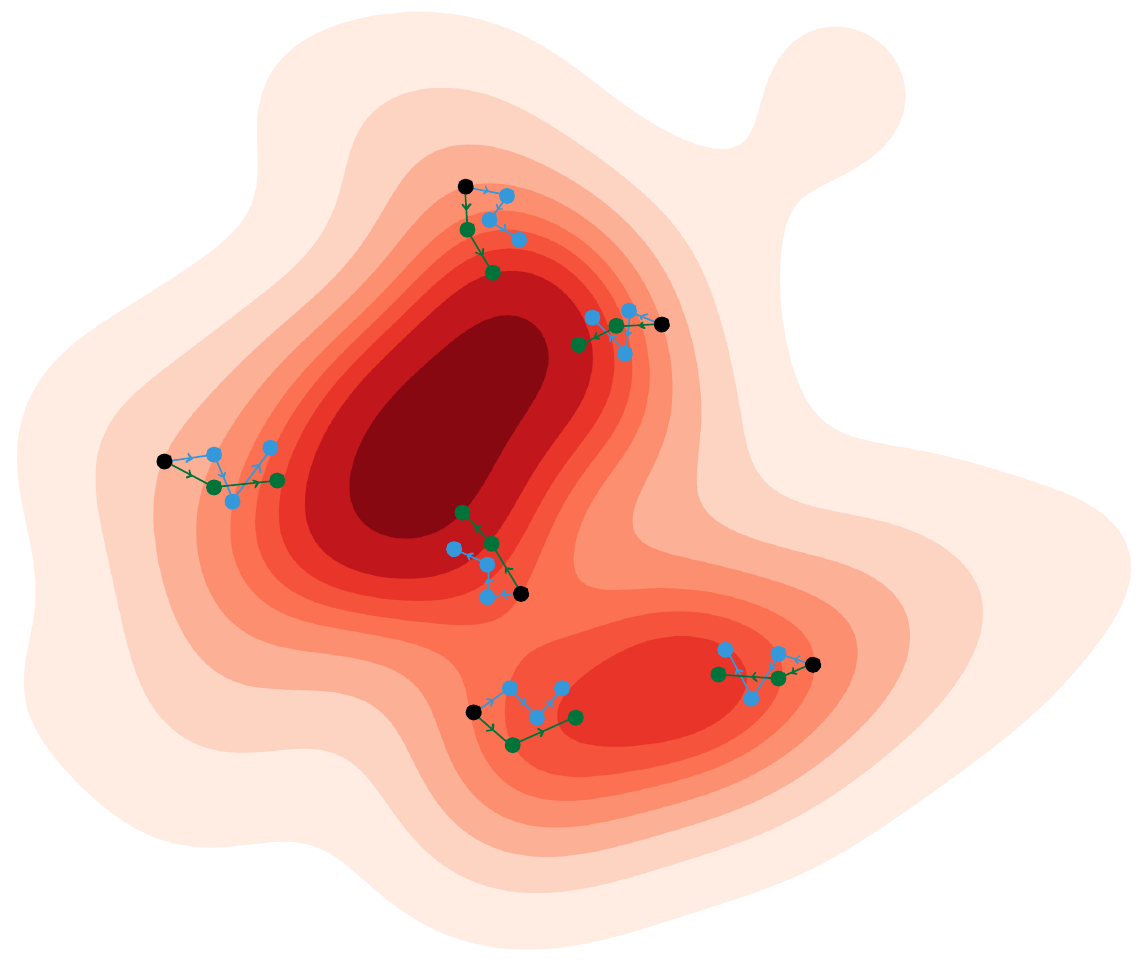}
    \end{minipage}
    \begin{minipage}[b]{0.35\textwidth}
        \centering
        \includegraphics[width=0.99\textwidth]{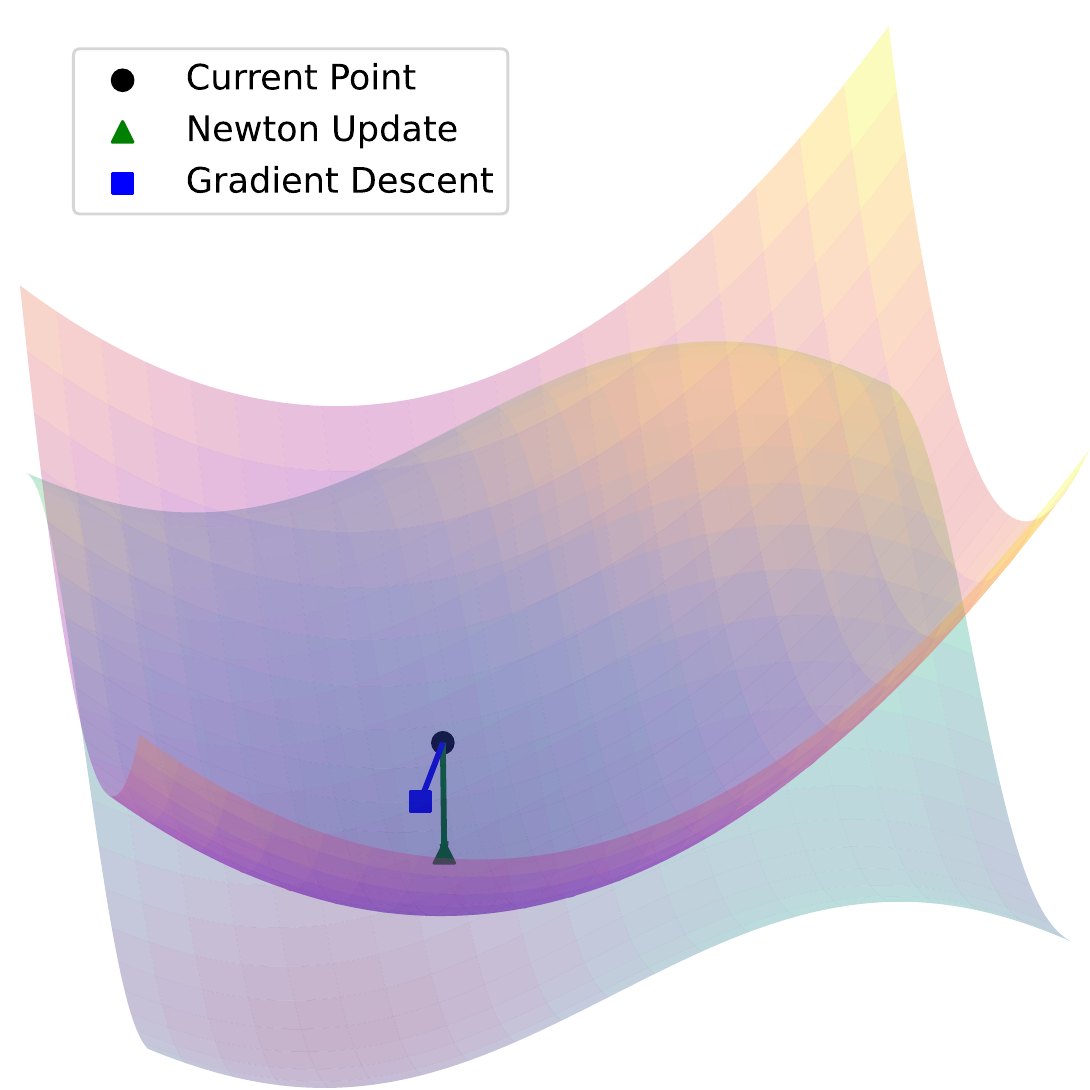}
    \end{minipage}
    \caption{Conceptual overview of the SVN method. The green curvature-informed SVN updates are much higher quality and require fewer steps than the corresponding blue SVGD ones.}
    \label{fig:Overview}
\end{figure}

The recent resurgence of interest in second-order methods in machine learning stems from more efficient Hessian approximations \citep{heskes2000natural, botev2017practical, martens2015optimizing} and the availability of plug-and-play software libraries \citep{dangel2020backpack, osawa2023asdl}. These advances have made Hessians more practical, particularly in the Laplace approximation for Bayesian deep learning \citep{mackay1992bayesian, laplace1774memoires}, now available as a PyTorch package \citep{daxberger2022laplace}. This progress has motivated us to re-evaluate the use of Hessians in ensemble-based Bayesian inference methods.

The primary contribution of this work is demonstrating SVN's capabilities in BNNs through extensive empirical studies and providing a user-friendly guide and modular codebase for practical deployment. Our adaptation, Stein Variational Newton Neural Network Ensembles, incorporates local curvature information of the posterior landscape using modern, scalable Hessian approximations, achieving faster convergence than traditional gradient-based methods. Additionally, we use geometry-aware kernels within the SVN framework to enhance convergence and performance. \cref{fig:Overview} illustrates how SVN ensembles make more informed gradient updates by incorporating second-order information.

Our main contributions are to
\begin{itemize}
    \item Incorporate modern, scalable Hessian approximations into particle-based inference
    \item Extend SVN from simple low-dimensional problems to deep neural network posteriors, including with a scalable last-layer version
    \item Demonstrate effective BNN inference on various synthetic and real-world datasets
\end{itemize}

\section{Stein Variational Newton Neural Network Ensembles}
\label{sec:svn algo}

The SVN algorithm \citep{detommaso2018stein} enhances SVGD \citep{liu2016stein} by incorporating curvature or Hessian information. While Hessian approximations are more computationally involved, they significantly improve convergence rates and stability by leveraging second-order optimization principles. We will first discuss these techniques in deep learning, focusing on Newton's method. Next, we will explore various approaches for modern Hessian approximations. Finally, we will introduce the Stein Variational Newton (SVN) algorithm. Consider a dataset $\mathcal{D} = \{(x_j \in \mathbb{R}^m, y_j \in \mathbb{R}^k) \}_{j=1}^n$ and a neural network model $g_\phi: \mathbb{R}^m \rightarrow \mathbb{R}^k$, parameterized by $\phi$. Let $\ell: \mathbb{R}^k \times \mathbb{R}^k \rightarrow \mathbb{R}$ be the loss function, e.g., mean squared error, or negative log-likelihood. The empirical risk for a batch of data $\mathcal{B} = \{(x_i \in \mathbb{R}^m, y_i \in \mathbb{R}^k) \}_{i=1}^b \subset \mathcal{D}$, with $|\mathcal{B}| = b$, is defined as:

\begin{align}
\mathcal{L}(\mathcal{B}; \phi) = \mathbb{E}_{\tilde{y} \sim p(y | g_\phi(x))} \left[ \ell(g_\phi(x), \tilde{y}) \right] \approx  \frac{1}{b}  \sum_{i=1}^{b} \ell(g_\phi(x_i), y_i).    
\label{eq:empirical_risk}    
\end{align}

Newton's method optimizes the empirical risk using a second-order Taylor approximation:

\begin{align}
\mathcal{L}(\mathcal{B}; \phi) \approx \mathcal{L}(\mathcal{B}; \phi_t) 
+ (\phi - \phi_t)^\top \nabla_\phi \, \mathcal{L}(\mathcal{B}; \phi_t) + \frac{1}{2}(\phi - \phi_t)^\top \mathbf{H}_{\phi_t} (\phi - \phi_t),
\label{eq:second_order_taylor}    
\end{align}

at timestep $t$, where

\begin{equation}
\mathbf{H}_{\phi_t} = \nabla^2_\phi \mathcal{L}(\mathcal{B}; \phi_t) \approx \frac{1}{b}  \sum_{i=1}^{b} \nabla^2_\phi \ell(g_\phi(x_i), y_i),
\label{eq:hessian def}
\end{equation}

is the Hessian matrix of $\mathcal{L}$ with respect to $\phi$, evaluated at $\phi_t$ for a given batch $\mathcal{B}$. Following \citet{Goodfellow-et-al-2016}, the Newton parameter update would then be defined as:

\begin{equation}
\phi_{t+1} = \phi_t - \mathbf{H}_{\phi_t}^{-1} \nabla_\phi \mathcal{L}(\mathcal{B}; \phi_t).     
\label{eq: newton update}
\end{equation}

This update step adjusts parameters by considering both the gradient and curvature of the empirical risk function. For locally quadratic functions with a positive definite Hessian, rescaling the gradient by \( \mathbf{H}^{-1} \) allows Newton's method to move directly to the minimum. For convex functions with higher-order terms, iterative updates lead to faster and more stable convergence than first-order methods, a feature that we visualize in \cref{fig:Overview} and demonstrate empirically for SVN Ensembles in \cref{sec: Experiments}.

\subsection{Hessian Approximations in Deep Learning}
\label{subsec: hessian approximations}

As previously noted, a crucial component in implementing the SVN algorithm for neural networks is the Hessian matrix $\mathbf{H}_{\phi}$ \citep{detommaso2018stein}. However, directly computing the Hessian, as defined in \cref{eq:hessian def}, is often impractical due to its quadratic scaling with the number of network parameters. Therefore, a good approximation is necessary. A positive-definite approximation is particularly beneficial for optimization purposes, as mentioned above. In this context, we will consider the Fisher Information Matrix (FIM) \citep{amari1998natural} defined for batch $\mathcal{B}$ as follows:

\begin{equation}
\mathbf{F}_\phi = \,  \sum_{i=1}^{b} \mathbb{E}_{\tilde{y} \sim p(y | g_\phi(x_i))} \left[ J_\phi(\tilde{y} | x_i) \, J_\phi(\tilde{y} | x_i)^\top \, \right],
\label{eq: fisher}
\end{equation}

with $J_\phi(\tilde{y} | x_i) = \nabla_\phi \log p(\tilde{y}|g_\phi(x_i)) $. It is also possible to use the Generalized-Gauss-Netwon (GGN) approximation \citep{Schraudolph_ggn}: 

\begin{equation}
\mathbf{G}_\phi = \sum_{i=1}^{b} \, \nabla_\phi g_\phi(x_i) \left( \nabla^2_\phi \log p(y_i | g_\phi(x_i)) \right)  \nabla_\phi g_\phi(x_i)^\top.     
\label{eq: ggn}
\end{equation}

\begin{figure}
    \centering
    \includegraphics[width=0.8\linewidth]{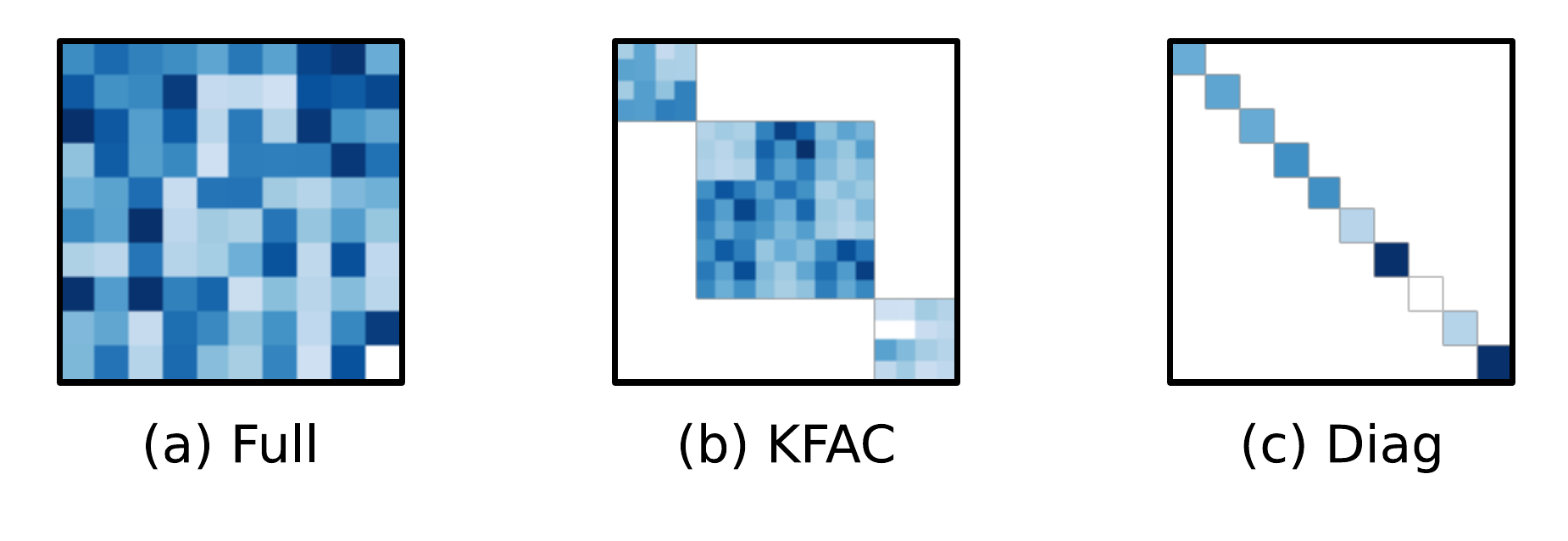}
    \caption{Overview of the Hessian approximations used in our SVN algorithm.}
    \label{fig:hessian approx overview}
\end{figure}

For most common likelihoods, and the ones we consider in this paper, the FIM and GGN are equivalent \citep{martens2020new}. 
For the deep Laplace approximation, these have emerged as the default choices \citep{ritter2018scalable, ritter2018online, kristiadi2020being, lee2020estimating, daxberger2021bayesian, immer2021improving} and are thus the ones we implement in this work. 
While $\mathbf{F}$ and $\mathbf{G}$ can be estimated efficiently, their entries are still quadratic in the number of parameters.
We thus consider three Hessian approximations in our work. \textbf{Full} represents the matrices as defined in \cref{eq: fisher} and \cref{eq: ggn}. The \textbf{Diagonal} \citep{denker1990transforming, lecun1990optimal} approximation considers only diagonal entries of these matrices.
Lastly, the Kronecker-Factored Approximate Curvature \citep[\textbf{KFAC} or \textbf{Kron};][]{heskes2000natural, martens2015optimizing, botev2017practical} considers block-diagonal entries that correspond to the layers of the neural network. These are represented as lightweight Kronecker factors. The KFAC can be improved by using low-rank approximation factors~\citep{lee2020estimating} and leveraging the eigendecomposition \citep{kfac_eigendecomp}. An overview of these methods is shown in \cref{fig:hessian approx overview}.
They are efficiently implemented for NNs by \citet{daxberger2022laplace}.

\subsection{Particle-based BNN inference}

The goal of BNN inference is to approximate the intractable Bayesian posterior distribution $p(\phi \mid \mathcal{D})$ on $\mathbb{R}^d$, abbreviated as $\pi$, where $d$ is the dimensionality of $\phi$ and number of parameters in the neural network $g_\phi$. In particle-based BNN inference, $\pi$ is approximated using an ensemble of $N$ models $\{g_{\phi_1}, \cdots, g_{\phi_N} \}$, which can be written as:

\begin{equation}
\rho(\phi) = \frac{1}{N} \sum^N_{i=1} \delta_{\phi_i}(\phi),
\end{equation}

where $\delta_\phi(x)$ denotes the Dirac delta function centered at $\phi$. In \textbf{Deep Ensembles} \citep{lakshminarayanan2017simple}, each neural network is trained independently with MAP training. Due to the stochasticity of neural network training, $g_{\phi_i} \ne g_{\phi_j}$, for $i \ne j$. 
A major breakthrough in approximate Bayesian inference was \textbf{Stein Variational Gradient Descent} \citep{liu2016stein}, which defines a sequence of transport maps \( T = T_1 \circ \cdots \circ T_k \circ \cdots \) to minimize \(\mathcal{D}_\text{KL} (T(\rho) \| \pi)\). This method updates the particles at each iteration to minimize the Kullback-Leibler (KL) divergence \citep{kl_divergence} between the true posterior \(\pi\) and our approximation \(T(\rho)\).
The update rule $T_l$ is chosen to be a perturbation of the identity mapping in the vector-valued RKHS $\mathcal{H}^d = \mathcal{H} \times \cdots \times \mathcal{H}$. The sequence of maps $\{T_1, T_2, \cdots, T_l, \cdots \}$ then corresponds to gradient descent with respect to the KL divergence. Following \citet{liu2016stein}, $T_l$, with $\phi_i^{l+1} = T_l\phi_i^l$, can be expressed explicitly as: $T_{l} = I - \epsilon \, v^{\operatorname{SVGD}}_l(\cdot)$, where

\begin{align}
v^{\operatorname{SVGD}}_l(\phi) = \frac{1}{N} \sum^N_{j=1}  &\underbrace{k(\phi_j^l, \phi) \, \nabla_{\phi_j^l}  \log \pi (\phi_j^l)}_{\text{weighted average steepest descent direction}} + \underbrace{\nabla_{\phi_j^l} k(\phi_j^l, \phi)}_{\text{repulsive force}} 
\label{eq: svgd operator}
\end{align}

Since $I$ is the identity mapping, $ v^{\operatorname{SVGD}}_l(\cdot) $ serves as a gradient update. The first term moves particles towards high-probability regions in $\pi$, while the second term acts as a 'repulsive force,' dispersing particles to prevent clustering around a single mode. SVGD was extended to neural network ensembles by \citet{d2021stein}, and we will compare our results to this method in \cref{sec: Experiments}.

\subsection{Stein Variational Newton}
\label{subsec: SVN}

The SVN algorithm, as proposed by \citet{detommaso2018stein}, enhances SVGD by performing steepest gradient descent in the functional Newton direction. This is represented by the transport map  $T_{l} = I - \epsilon \, v^{\operatorname{SVN}}_l(\cdot)$ at time step $l$. The SVN update can be expressed as a Galerkin approximation in the finite-dimensional linear space \(\mathcal{H}_N^d = \text{span} \{ k(\phi_1, \cdot), \cdots, k(\phi_N, \cdot) \}\). The $j$-th component of $ v^{\operatorname{SVN}}$ then corresponds to:

\begin{equation}
v^{\text{SVN}}_j(\phi) = \sum^N_{k=1} \alpha_j^k \, k(\phi_k, \phi)
\label{eq: svn component}
\end{equation}

We use the Hessians $\mathbf{H}_{\phi_1}, \cdots, \mathbf{H}_{\phi_N}$ to solve for the alpha vector $\vec{\alpha}_j = (\alpha_j^1, \cdots, \alpha_j^N)^{\top}$ for each particle $\phi_j$. Following the framework in \citet{gallego2020stochastic} and \citet{leviyev2022stochastic}'s SVN formulation, we define the kernel matrix $K \in \mathbb{R}^{Nd \times Nd}$ as:

\begin{equation}
K = \frac{1}{N}\left(\begin{array}{ccc}
k\left(\phi_1, \phi_1\right) I_{d \times d} & \cdots & k\left(\phi_1, \phi_N\right) I_{d \times d} \\
\vdots & \ddots & \vdots \\
k\left(\phi_N, \phi_1\right) I_{d \times d} & \cdots & k\left(\phi_N, \phi_N\right) I_{d \times d}
\end{array}\right),    
\label{eq: K matrix}
\end{equation}

where $N$ refers to the number of particles as before, and $d$ is the number of parameters in each ensemble member. Therefore, a vector of length $Nd$ can represent the whole ensemble. Furthermore, for $1 \leq m, n \leq N$, we define the matrix blocks $h^{m,n} \in \mathbb{R}^{d \times d}$,

\begin{align}
h^{m, n} &:= \frac{1}{N} \sum_{p=1}^N \left[-k\left(\phi_p, \phi_m\right) k\left(\phi_p, \phi_n\right) \mathbf{H}_{\phi_p} \right. + \left. \nabla_{\phi_p} k\left(\phi_p, \phi_n\right) \, \nabla_{\phi_p} k\left(\phi_p, \phi_m\right)^\top \right].
\label{eq: matri blocks svn hessian}
\end{align}

The first term in the equation above can be regarded as a weighted kernel average of the Hessians in \cref{eq: newton update}. The second term corresponds to a repulsive force between particles in the RKHS similar to \cref{eq: svgd operator}. Denoting the SVN-Hessian as
\begin{equation}
H^{\operatorname{SVN}} =\left[\begin{array}{ccc}
h^{11} & \cdots & h^{1 N} \\
\vdots & \ddots & \vdots \\
h^{N 1} & \cdots & h^{N N}
\end{array}\right],    
\label{eq: svn hessian}
\end{equation}

computing the SVN update amounts to solving the following linear system of $Nd$ equations:

\begin{equation}
H^{\operatorname{SVN}} \alpha=v^{\operatorname{SVGD}}.
\label{eq: linear system}    
\end{equation}

Finally, solving for $\alpha \in \mathbb{R}^{Nd}$, the SVN update vector corresponding to the functional Newton direction of steepest descent of the KL divergence can be calculated as:

\begin{equation}
v^{\operatorname{SVN}}=N K \alpha.
\label{eq: svn update computation}
\end{equation}

\subsection{SVN Ensembles Algorithm}

\begin{figure*}[h]
    \centering
    \begin{subfigure}[b]{0.32\linewidth}
        \centering
        \includegraphics[width=\textwidth]{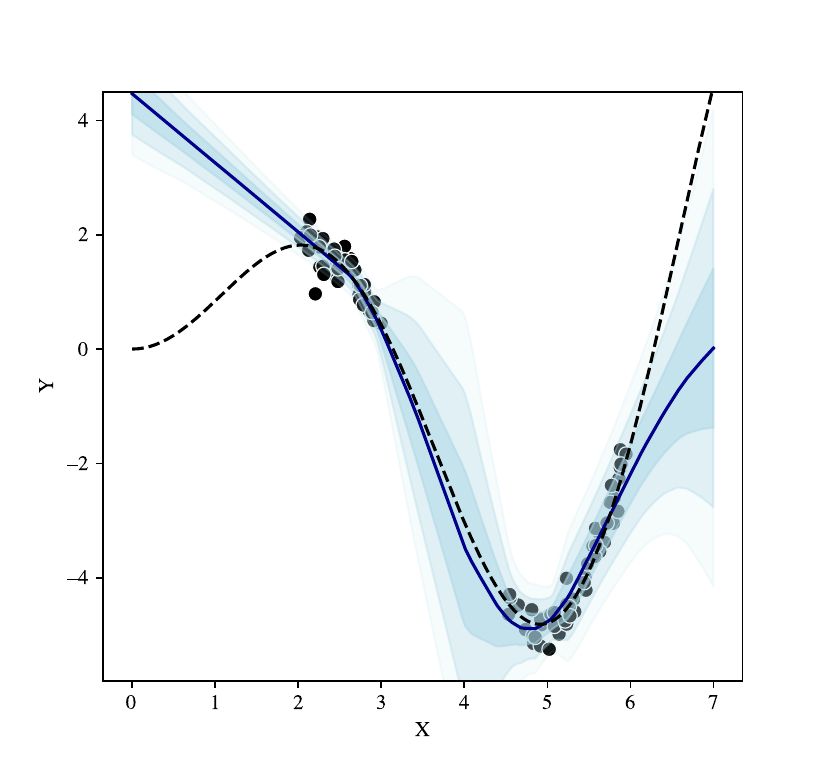}
        \caption*{Ensemble}
    \end{subfigure}
    \begin{subfigure}[b]{0.32\linewidth}
        \centering
        \includegraphics[width=\textwidth]{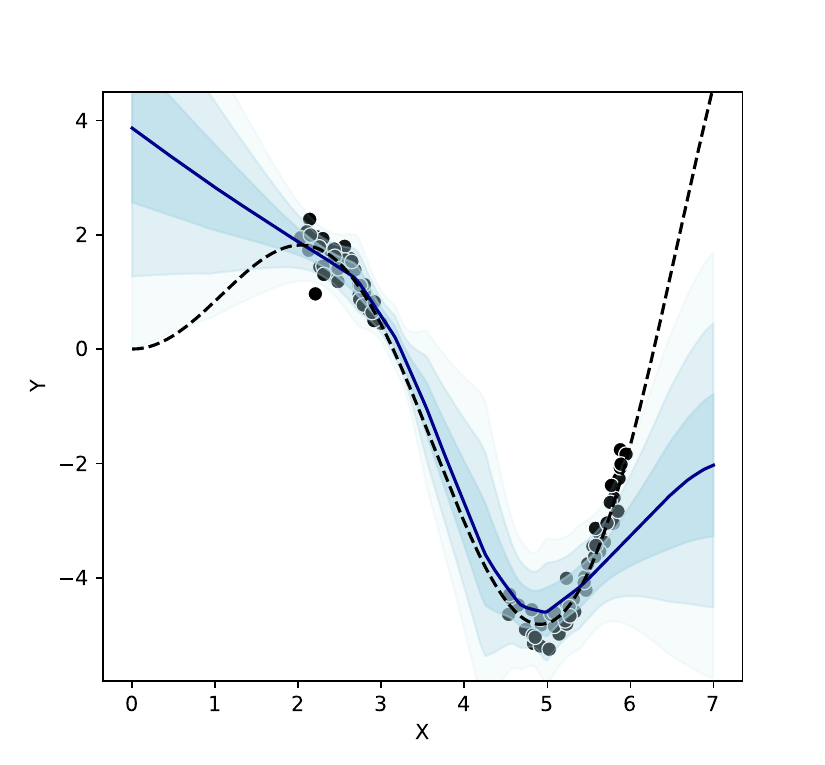}
        \caption*{SVGD}
    \end{subfigure}
    \begin{subfigure}[b]{0.32\linewidth}
        \centering
        \includegraphics[width=\textwidth]{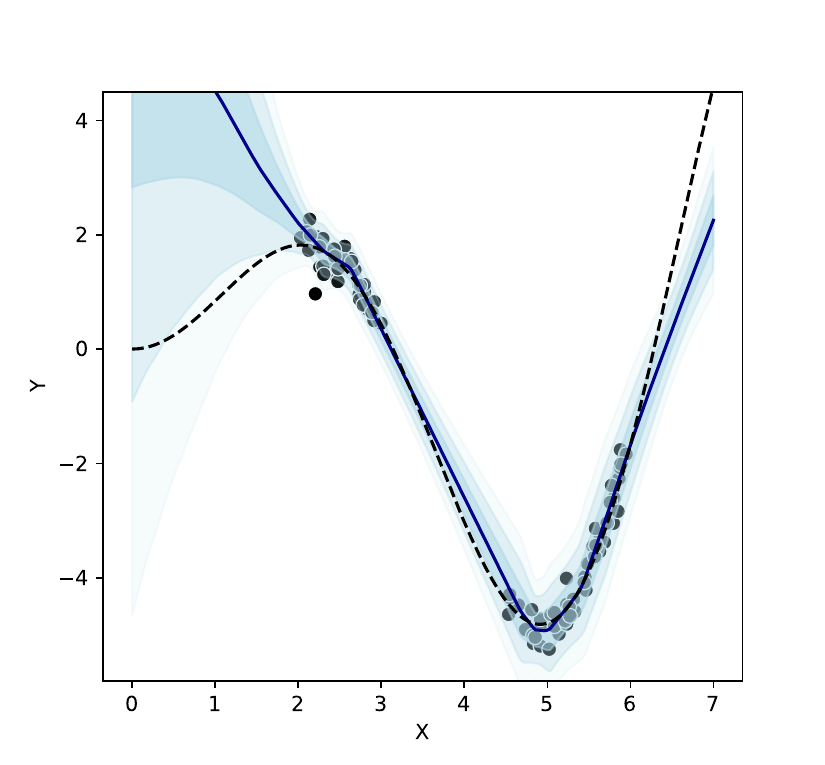}
        \caption*{SVN}
    \end{subfigure}
    \caption{Synthetic regression example for Ensemble, SVGD, and SVN methods. The training data is marked with black dots, and the true function is represented with a dashed line. The predictive mean of the neural network ensemble is shown in dark blue, with the standard deviations highlighted in light blue. SVN best captures the underlying data distribution.}
    \label{fig:toy_regression}
\end{figure*}

\cref{alg: SVN algorithm} summarizes SVN neural network ensembles. While we demonstrate in \cref{sec: Experiments} that the method above is computationally feasible and yields better results for moderately-sized neural networks with the full Hessian approximation, the linear system in \cref{eq: linear system} has $Nd$ equations and needs to be computed for every gradient step. Moreover, a na\"ive implementation of $K$ scales with $\mathcal{O}(N^2 d^2)$ in memory terms, which is extremely prohibitive for larger deep learning models.

To this end, \citet{detommaso2018stein} already proposed two modifications to the above procedure to improve its scalability. The first approach is using the Newton-conjugate-gradient (NCG) \citep[ Chapters 5 and 7]{Wright1999NumericalOptimization}, which approximates \cref{eq: linear system}.
Furthermore, this approach only requires evaluating the matrix-vector product $H^{\operatorname{SVN}} \alpha$, without explicitly constructing the matrix. This allows us to efficiently use the KFAC and Diagonal approximations detailed in \cref{subsec: hessian approximations}. Additionally, one can consider a block diagonal approximation to \cref{eq: linear system}, where the off-diagonal matrix blocks in \cref{eq: svn hessian} are disregarded, and we solve the following linear system for each particle:

\begin{equation}
h^{m,m} \alpha_m = v^{\text{SVGD}}_m \quad \text{ for } \quad m=1, \cdots, N. 
\label{eq: block diag linear system}
\end{equation}

This computes $\alpha_m$ for each particle independently and involves only \(d\) equations. Applying the inverse of the SVN Hessian matrix block $h^{m,m}$ to both sides of \cref{eq: block diag linear system} makes the computation of the $\alpha$ vector in kernel space similar to the Newtonian parameter update vector discussed in \cref{eq: newton update} at the beginning of this section.

\begin{algorithm}[t]
\caption{One iteration of our SVN neural network ensemble algorithm}
\begin{algorithmic}[1]
\STATE \textbf{Input:} Particles $\{\phi_i^l\}_{i=1}^N$ at stage $l$; step size $\varepsilon$, kernel $k(\cdot, \cdot) \in \{ k^{M_{\text{SVN}}}, k^I \}$, Hessian approximation $\mathbf{H} \in \{\mathbf{H}_{\text{Full}}, \mathbf{H}_{\text{KFAC}}, \mathbf{H}_{\text{Diag}} \}$
\STATE \textbf{Output:} Particles $\{\phi_i^{l+1}\}_{i=1}^N$ at stage $l+1$
\STATE Compute $v^{\operatorname{SVGD}}_l$
\FOR{$i=1,2,\ldots, n$}
    \IF{Block Diagonal Approximation}
        \STATE Solve the linear system from \cref{eq: block diag linear system} for $\alpha^1, \ldots, \alpha^n$
    \ELSE
        \STATE Solve the linear system from \cref{eq: linear system} for $\alpha^1, \ldots, \alpha^n$
    \ENDIF
    \STATE Set $\phi_i^{k+1} \leftarrow \phi_i^k + \varepsilon \, v_l^{\operatorname{SVN}}\left(\phi_i^{k} \right)$ given $\alpha^1, \ldots, \alpha^n$
\ENDFOR
\end{algorithmic}
\label{alg: SVN algorithm}
\end{algorithm}

\paragraph{Hessian kernel.}
Our Hessian approximation $\mathbf{H}_\phi \approx \nabla^2 \log \pi$ characterizes the local curvature of the posterior.
Following \citet{detommaso2018stein}, we consider the average curvature metric of the posterior $\pi$ with respect to our ensemble: 

\begin{equation}
M_\pi = \mathbb{E}_{\phi \sim \pi} [\nabla^2_\phi \log \pi (\phi) ] \approx \frac{1}{N} \sum^N_{j=1} \mathbf{H}_{\phi_j} = M_{SVN},
\label{eq: average curvature matrix}    
\end{equation}

and use it to construct an anisotropic Gaussian Kernel \citep{liu2016stein}:

\begin{equation}
k^M_l(\phi, \phi') := \exp \left( - \frac{1}{2d} \| \phi - \phi' \|_{M_{\text{SVN}} } \right),
\label{eq: anistrpoic gaussian kernel}
\end{equation}

with $\| \phi \|_{M_{\text{SVN}}} = \phi^\top \, M_{\text{SVN}} \, \phi$. In \cref{app sub sec: ablation}, we perform ablation studies between using $M = M_{\text{SVN}}$ and simply using the scaled Gaussian kernel with $M=I$.

\paragraph{LL-SVN.}
Inspired by recent work on subnetwork inference \citep{daxberger2021bayesian} and competitive performance of the last-layer Laplace approximation \citep{daxberger2022laplace}, we propose a novel and computationally efficient modification to \cref{alg: SVN algorithm}, called \textbf{LL-SVN}, where the SVN update $v^{\operatorname{SVN}}$ is performed only for the last layer of the network, while $v^{\operatorname{SVGD}}$ is used for the other layers.
This approach leverages high-quality curvature information in the most expressive part of the neural network, better utilizing the learned representation. We compute $v^{\operatorname{SVGD}}_l$ for the entire network, then use the last $d_{ll}$ elements, corresponding to the last layer, to compute the SVN vector $v^{\operatorname{SVGD}}_l$. \Cref{alg: LL-SVN algorithm} in \Cref{app: SVN algo details} summarizes this procedure.

\section{Experiments}
\label{sec: Experiments}

We evaluate the proposed Stein Variational Newton neural network ensemble (\textbf{SVN}) on a diverse range of synthetic and real-world datasets, demonstrating its competitive or superior performance for BNN inference. Our approach is benchmarked against other particle-based VI methods, including deep ensembles (\textbf{Ensemble}) \citep{lakshminarayanan2017simple}, repulsive ensembles \citep{dangelo2023repulsive} (specifically, the weight-space version (\textbf{WGD})), and Stein Variational Gradient Descent (\textbf{SVGD}) \citep{liu2016stein} for neural network ensembles \citep{d2021stein}. All these methods involve different training schemes for a collection of neural networks. Thus, the resulting ensemble model, denoted as $\{ g_{\phi_1}, \cdots, g_{\phi_N} \}$, can be uniformly evaluated across all methods.
Note, however, that deep ensembles do not asymptotically converge to the true posterior as $N \rightarrow \infty$.

\begin{figure*}[h]
    \centering
    \includegraphics[width = 0.9\linewidth]{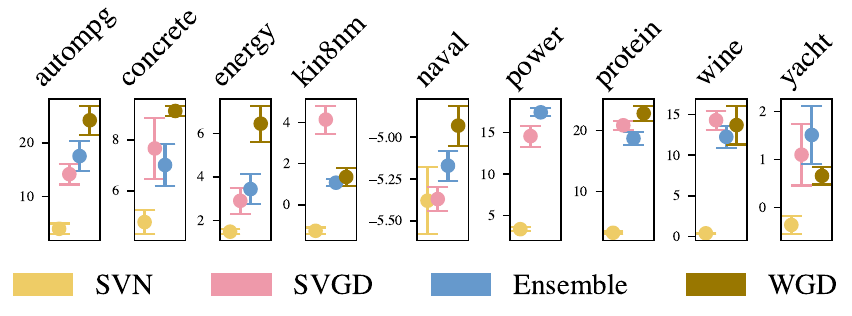}
    \caption{Test negative log-likelihood on UCI regression datasets. We truncated the power plot as a result of WGD's inferior performance. Our proposed SVN method outperforms the ensemble, WGD, and SVGD on all datasets except for naval.}
    \label{fig: uci regression NLL}
\end{figure*}

\subsection{Toy Regression}

As a sanity check, we start with a synthetic one-dimensional regression problem using a self-generated dataset. Details on data generation are in \cref{app sub sec: toy regression}. In \cref{fig:toy_regression}, we plot the predictive mean of the ensemble, $\Bar{g} = \frac{1}{N} \sum_{i=1}^N g_{\phi_i}$, along with the predictive mean plus or minus three standard deviations in increasingly lighter shades of blue. Training data points are in black, and the true function is represented with a dashed line. The plots show that while the ensemble and SVGD yield better uncertainties between the two clusters, they are underconfident within the clusters and overconfident outside the data region. In contrast, SVN's predictions and uncertainties best match the underlying function and given data, capturing the data distribution most effectively.

\subsection{UCI Benchmark}

To demonstrate the effectiveness of SVN on real-world datasets, we follow \citet{hernándezlobato2015probabilistic} and consider several regression and binary classification datasets from the UCI Machine Learning Repository \citep{dua2017uci}. We use 5-fold cross-validation, combining the training and test data. Additionally, we split off $20\%$ of the training data for validation. We report the mean performance metrics along with the standard error. The primary metric considered is the Gaussian negative log-likelihood (NLL) \citep{gaussian_nll_loss} on the test dataset. In the regression tasks of \cref{fig: uci regression NLL}, SVN generally outperforms other methods, except for the naval dataset. The corresponding MSE and NLL results are listed in \cref{tab: uci regression NLL} and \cref{tab: uci regression MSE} in the Appendix. Next, we evaluate SVN on the UCI binary classification tasks using standard metrics: accuracy, NLL, and AUROC. We also include calibration metrics such as the Expected Calibration Error (ECE) \citep{guo2017calibration} and the Brier score \citep{brier_Score}, which has been argued to offer many benefits over the ECE \citep{gruber2024better}. As shown in \cref{tab: uci Classification}, SVN performs competitively across all five tasks, outperforming the baselines on three datasets with respect to NLL. Further details on this experiment can be found in \cref{app sub sec: uci experiments}.

\begin{table}[h]
\caption{Classification metrics on test datasets of the UCI Binary Classification tasks. Our method is competitive on all datasets and outperforms the baselines on three datasets in terms of NLL.}
\label{tab: uci Classification}
\centering
\begin{tabular}{ccccccc}
\toprule
Dataset & Methods & ACC $\uparrow$ & NLL $\downarrow$   & ECE $\downarrow$   & Brier $\downarrow$ & AUROC $\uparrow$ \\
\midrule
\multirow{5}{*}{australian } 
 & Ensemble & $\mathbf{0.84_{0.01}}$ & $1.04_{0.12}$ &  $\mathbf{0.17_{0.01}}$ & $\mathbf{0.24_{0.01}}$ & $\mathbf{0.91_{0.01}}$ \\
 & WGD & $\mathbf{0.84_{0.01}}$ & $1.05_{0.10}$ &  $\mathbf{0.16_{0.01}}$ & $\mathbf{0.24_{0.02}}$ & $\mathbf{0.92_{0.01}}$ \\
 & SVGD & $\mathbf{0.84_{0.02}}$ & $1.03_{0.10}$ &  $\mathbf{0.15_{0.01}}$ & $\mathbf{0.24_{0.02}}$ & $\mathbf{0.92_{0.01}}$ \\
 & SVN & $\mathbf{0.84_{0.01}}$ & $\mathbf{0.62_{0.06}}$  & $\mathbf{0.15_{0.01}}$ & $\mathbf{0.26_{0.01}}$ & $\mathbf{0.90_{0.01}}$ \\ \bottomrule

\multirow{5}{*}{breast} 
 & Ensemble & $\mathbf{0.97_{0.00}}$ & $0.23_{0.05}$ &  $\mathbf{0.03_{0.00}}$ & $\mathbf{0.05_{0.00}}$ & $\mathbf{0.99_{0.00}}$ \\
 & WGD & $0.96_{0.00}$ & $0.29_{0.08}$ &  $0.04_{0.00}$ & $\mathbf{0.05_{0.00}}$ & $0.96_{0.02}$ \\
 & SVGD & $\mathbf{0.97_{0.00}}$ & $0.38_{0.13}$  & $\mathbf{0.03_{0.00}}$ & $\mathbf{0.05_{0.00}}$ & $0.98_{0.00}$ \\
 & SVN & $\mathbf{0.97_{0.00}}$ & $\mathbf{0.13_{0.02}}$ &  $0.04_{0.00}$ & $\mathbf{0.05_{0.01}}$ & $0.97_{0.00}$ \\ \bottomrule

\multirow{5}{*}{ionosphere} 
 & Ensemble & $\mathbf{0.86_{0.02}}$ & $\mathbf{0.40_{0.04}}$ &  $\mathbf{0.14_{0.01}}$ & $\mathbf{0.21_{0.02}}$ & $\mathbf{0.94_{0.01}}$ \\
 & WGD & $0.82_{0.01}$ & $1.31_{0.11}$ &  $0.18_{0.01}$ & $0.28_{0.01}$ & $\mathbf{0.92_{0.01}}$ \\
 & SVGD & $\mathbf{0.86_{0.01}}$ & $\mathbf{0.42_{0.04}}$  & $\mathbf{0.15_{0.01}}$ & $\mathbf{0.21_{0.02}}$ & $\mathbf{0.92_{0.01}}$ \\
 & SVN & $\mathbf{0.86_{0.01}}$ & $\mathbf{0.40_{0.04}}$ & $\mathbf{0.16_{0.01}}$ & $\mathbf{0.21_{0.02}}$ & $\mathbf{0.93_{0.01}}$ \\\bottomrule
\multirow{5}{*}{parkinsons} 
 & Ensemble & $0.85_{0.03}$ & $\mathbf{0.43_{0.04}}$  & $0.18_{0.01}$ & $0.26_{0.03}$ & $0.85_{0.03}$ \\
 & WGD & $\mathbf{0.91_{0.03}}$ & $\mathbf{0.56_{0.27}}$  & $\mathbf{0.09_{0.03}}$ & $\mathbf{0.14_{0.05}}$ & $\mathbf{0.92_{0.03}}$ \\
 & SVGD & $\mathbf{0.92_{0.03}}$ & $\mathbf{0.47_{0.21}}$  & $\mathbf{0.09_{0.03}}$ & $\mathbf{0.14_{0.05}}$ & $\mathbf{0.97_{0.02}}$ \\
 & SVN & $\mathbf{0.91_{0.03}}$ & $\mathbf{0.29_{0.10}}$  & $\mathbf{0.11_{0.02}}$  & $\mathbf{0.15_{0.05}}$ & $\mathbf{0.94_{0.03}}$ \\ \bottomrule

\multirow{5}{*}{heart} 
 & Ensemble & $\mathbf{0.78_{0.02}}$ & $1.22_{0.09}$ &  $\mathbf{0.21_{0.01}}$ & $\mathbf{0.37_{0.02}}$ & $\mathbf{0.86_{0.01}}$ \\
 & WGD & $0.76_{0.01}$ & $2.02_{0.19}$  & $\mathbf{0.22_{0.01}}$ & $\mathbf{0.38_{0.02}}$ & $\mathbf{0.87_{0.01}}$ \\
 & SVGD & $\mathbf{0.79_{0.01}}$ & $2.14_{0.22}$  & $\mathbf{0.21_{0.01}}$ & $\mathbf{0.36_{0.01}}$ & $\mathbf{0.86_{0.01}}$ \\
 & SVN & $\mathbf{0.77_{0.02}}$ & $\mathbf{0.75_{0.03}}$  & $\mathbf{0.21_{0.01}}$  & $\mathbf{0.36_{0.02}}$ & $\mathbf{0.86_{0.01}}$ \\
\bottomrule
\end{tabular}
\end{table}

\begin{figure*}
    \centering
    \begin{subfigure}[b]{0.32\linewidth}
        \centering
        \includegraphics[width=\textwidth]{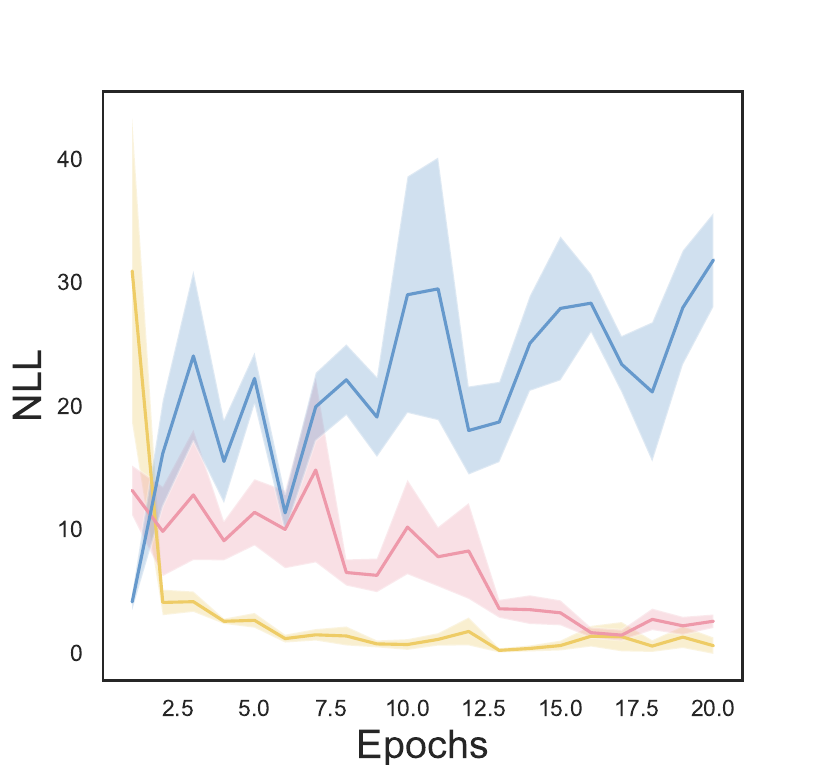}
        \caption*{Yacht}
    \end{subfigure}
    \begin{subfigure}[b]{0.32\linewidth}
        \centering
        \includegraphics[width=\textwidth]{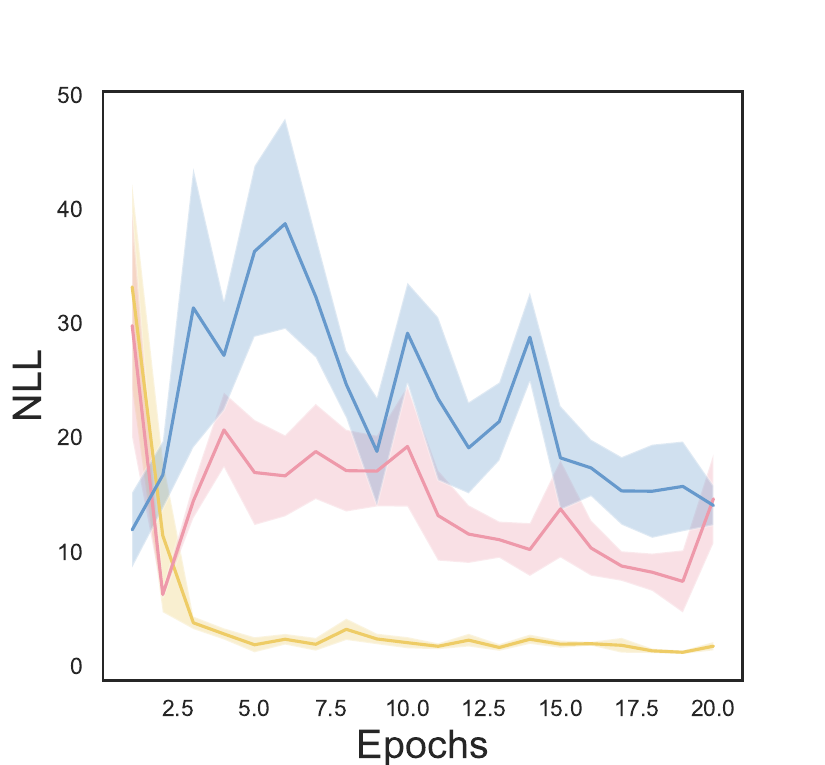}
        \caption*{Energy}
    \end{subfigure}
     \begin{subfigure}[b]{0.32\linewidth}
        \centering
        \includegraphics[width=\textwidth]{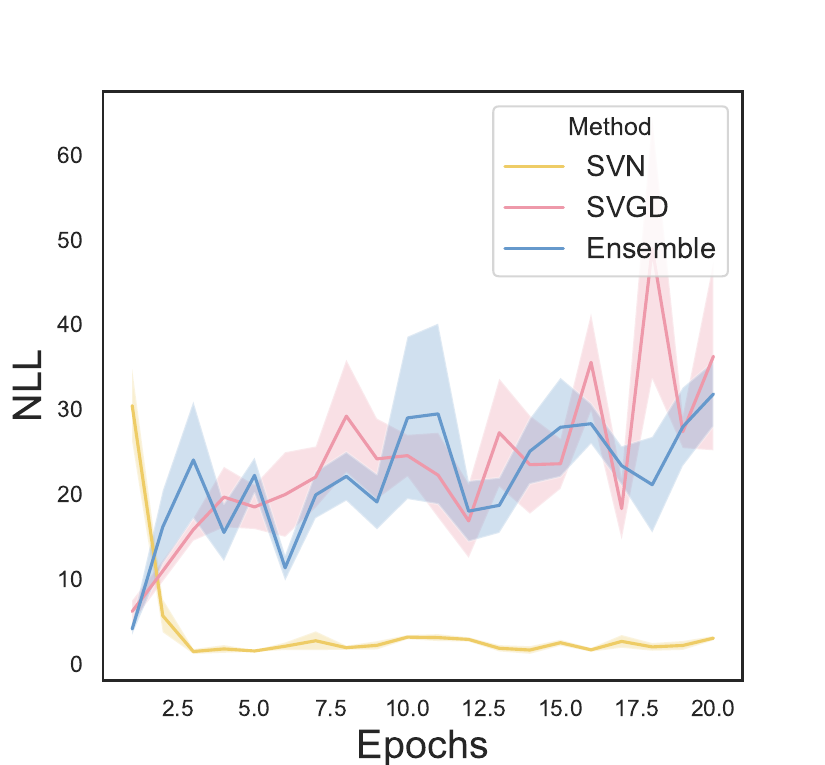}
        \caption*{Wine}
    \end{subfigure}
    \caption{Comparison of validation negative log-likelihood computed at the end of every epoch for the first $20$ epochs of training on Yacht, Energy, and Wine datasets. While SVN's initial performance is considerably worse than the other methods, it outperforms both within a few epochs.}
    \label{fig: convergence speed}
\end{figure*}

\subsection{Computer Vision Benchmark}

In order to further investigate our method, we now assess the capabilities of SVN on the computer vision benchmarks MNIST \citep{lecun1998mnist} and FashionMNIST \citep{xiao2017fashion}. Since these are multi-class classification problems, we use the evaluation metrics from the previous section. We use the LeNet \citep{lenet} architecture for our experiments. Since this network is considerably larger, we also evaluate our proposed \textbf{LL-SVN} algorithm. Recent work suggested that for sufficiently large neural networks \citep{farquhar2021liberty}, the diagonal approximation works well. As such, our SVN experiments are conducted with the diagonal approximation to the Hessian, while our LL-SVN uses the KFAC approximation to the last layer. Our results are detailed in \cref{tab: cv benchmark} and further experiment details can be found in \cref{app sub sec: cv experiments}.
We see that our methods are competitive with the baselines, and LL-SVN outperforms the baselines on FashionMNIST in terms of the NLL.

\begin{table}[htbp]
\caption{Computer vision image datasets. Our LL-SVN algorithm is competitive in almost all metrics and outperforms the ensemble, WGD, and SVGD on the more complicated FashionMNIST dataset in terms of the NLL.}
\label{tab: cv benchmark}
\centering
\begin{tabular}{ccccccc}
\toprule 
Dataset & Methods & ACC $\uparrow$ & NLL $\downarrow$  & ECE $\downarrow$   & Brier $\downarrow$ & AUROC $\uparrow$ \\
\midrule
\multirow{5}{*}{MNIST } 
 & Ensemble & $\mathbf{0.991_{0.000}}$ & $\mathbf{0.039_{0.001}}$  & $\mathbf{0.011_{0.000}}$ & $\mathbf{0.014_{0.000}}$ & $\mathbf{1.000_{0.00}}$ \\
 & WGD & $\mathbf{0.991_{0.000}}$ & $\mathbf{0.041_{0.003}}$  & $\mathbf{0.012_{0.001}}$ & $\mathbf{0.014_{0.001}}$ & $\mathbf{1.000_{0.000}}$ \\
 & SVGD & $\mathbf{0.991_{0.000}}$ & $\mathbf{0.038_{0.000}}$  & $\mathbf{0.012_{0.001}}$ & $\mathbf{0.014_{0.000}}$ & $\mathbf{1.000_{0.000}}$ \\
 & SVN & $\mathbf{0.991_{0.001}}$ & $0.041_{0.001}$  & $0.015_{0.001}$ & $0.016_{0.000}$ & $\mathbf{1.000_{0.000}}$ \\
 & LL-SVN & $\mathbf{0.991_{0.000}}$ & $\mathbf{0.038_{0.000}}$ & $\mathbf{0.011_{0.000}}$ & $\mathbf{0.014_{0.000}}$ & $\mathbf{1.000_{0.000}}$ \\
\midrule 
\multirow{5}{*}{FashionMNIST} 
 & Ensemble & $\mathbf{0.916_{0.001}}$ & $0.259_{0.005}$  & $\mathbf{0.051_{0.000}}$ & $\mathbf{0.122_{0.001}}$ & $\mathbf{0.995_{0.000}}$ \\
 & WGD & $0.912_{0.004}$ & $0.257_{0.009}$  & $\mathbf{0.056_{0.004}}$ & $\mathbf{0.128_{0.007}}$ & $\mathbf{0.994_{0.001}}$ \\
 & SVGD & $\mathbf{0.916_{0.002}}$ & $0.253_{0.006}$  & $\mathbf{0.051_{0.001}}$ & $\mathbf{0.123_{0.002}}$ & $\mathbf{0.995_{0.000}}$ \\
 & SVN & $0.907_{0.005}$ & $0.261_{0.009}$  & $0.0561_{0.001}$ & $0.135_{0.007}$ & $\mathbf{0.994_{0.001}}$ \\
 & LL-SVN & $\mathbf{0.918_{0.001}}$ & $\mathbf{0.241_{0.002}}$  & $\mathbf{0.051_{0.001}}$ & $\mathbf{0.120_{0.001}}$ & $\mathbf{0.995_{0.000}}$ \\
\bottomrule 
\end{tabular}
\end{table}

\subsection{Application to Intensive Care Unit Time Series Data}
\label{sub sec: icu dataset}

Following the task proposed in \citet{app13126930, winter2023predicting}, the results of our length of stay prediction study using the MIMIC-IV dataset \citep{johnson2024mimic} are presented in Table \ref{tab: mimic-iv benchmark}. We compared three methods: SVN, SVGD, and Ensemble. While the Ensemble method achieved the lowest MSE of 15.41 hours, it is crucial to note that in safety-critical domains such as healthcare, calibration is of paramount importance. In this regard, SVN demonstrates superior performance with the lowest NLL of 107.30. This indicates that SVN provides better-calibrated predictions, which is essential in clinical settings where accurate uncertainty quantification can significantly impact decision-making processes. The lower NLL suggests that SVN's predictions are not only reasonably accurate in terms of point estimates (as evidenced by its competitive MSE of 15.64 hours) but also offer more reliable probability distributions over the predicted length of stay. This balance of accuracy and well-calibrated uncertainty makes SVN particularly valuable for applications in critical care, where understanding the confidence in predictions is as important as the predictions themselves.

\begin{table}[htbp]
\caption{Length of stay prediction in ICU beds from the MIMIC-IV Dataset. The units are in hours. While the baselines are offering slightly better MSE, the calibration as exemplified by the NLL score is much better for our SVN method.}
\label{tab: mimic-iv benchmark}
\centering
\begin{tabular}{lcc}
\toprule
Method & MSE & NLL \\
\midrule
 Ensemble & $15.41_{0.07}$ & $438.84_{113.60}$ \\
 WGD & $15.49_{0.05}$ & $253.84_{10.90}$ \\
 SVGD & $\mathbf{15.22_{0.04}}$ & $226.62_{8.49}$ \\
 SVN & $15.64_{0.08}$ & $\mathbf{107.30_{41.81}}$ \\
\bottomrule
\end{tabular}
\end{table}

\subsection{Convergence Speed}
\label{sub sec: convergence speed}

As discussed in \cref{sec:svn algo}, incorporating curvature information in the optimization of deep neural networks can yield significantly faster convergence. We investigate whether this intuition holds for SVN ensembles. We visualize the validation NLL on three UCI datasets in \cref{fig: convergence speed} at the end of every epoch for the first $20$ epochs. Interestingly, SVN's initial updates are worse than SVGD, resulting in significantly higher validation NLL scores. However, within a few epochs, SVN surpasses the ensemble and SVGD quickly. We hypothesize that SVN requires a few iterations to fully leverage the curvature information in the posterior landscape to position its particles for more efficient Newtonian updates compared to the 'simple' SVGD updates. Moreover, we visualize some Hessians on a downscaled example in \cref{app sub sec: hessian plots}, showing that initially the Hessian is not very informative. As such, the SVN updates are likely not effective initially, and we only benefit from second-order optimization once our particles are better positioned in the posterior landscape.

\section{Related Work}
\label{sec:realted works}

\paragraph{BNN inference.}

Approximate BNN inference methods fall on a spectrum from cheap Laplace approximations \citep{laplace1774memoires, mackay1992bayesian, daxberger2022laplace, immer2021improving, immer2021scalable} and variational inference \citep[VI;][]{blei2017variational} to expensive Markov chain Monte Carlo \citep[MCMC;][]{andrieu2003introduction, welling2011bayesian} techniques, with a notable tradeoff in computational efficiency and estimation accuracy.
Arguably, approximations to the posterior can yield even better results than the actual posterior \citep{wenzel2020good}.
While there has been some recent interest in MCMC for deep neural networks \citep{izmailov2021bayesian, huang2023efficient, garriga2021exact}, the computational requirements of these techniques remain a challenge \citep{papamarkou2021challenges, izmailov2021bayesian, papamarkou2024position}.
However, mean-field VI, especially for small neural nets, has also recently been criticized \citep{farquhar2021liberty, foong2020expressiveness}.

\paragraph{Particle-based inference.}
Situated between mean-field VI and MCMC, particle-based VI approaches offer a good tradeoff between efficiency and performance \citep{liu2019understanding}.
Neural network ensembles have a rich history, beginning with the work of \citep{hansen90}. Deep ensembles \citep{lakshminarayanan2017simple} and variations \citep{ wenzel2020hyper, kobayashi2022disentangling, wild2023a} are relatively simple scalable methods.
Including repulsive forces in the particle updates even results in asymptotic convergence to the true posterior \citep{dangelo2023repulsive}.
Variants of SVGD \citep{liu2016stein} include projected SVGD \citep{chen2020projected_svgd}, Riemannian SVGD \citep{liu2017riemannian}, message-passing SVGD \citep{zhuo2018message}, annealed SVGD \citep{dangelo2021annealed}, as well as their extensions to NN Ensembles \citep{d2021stein}.
Variations of the SVN \citep{detommaso2018stein} include stochastic SVN \citep{leviyev2022stochastic}, which turns the method into an MCMC sampler by adding Gaussian noise, and projected SVN \citep{chen2020projected_svn}.
However, SVN has so far only been used in low-dimensional problems and not for high-dimensional BNN posteriors, as in our work.

\paragraph{Second-order methods.}
Second-order methods in VI have been pivotal in improving optimization efficiency, and parametric variational inference leveraging second-order information has shown accelerated convergence \citep{khan2017variational, khan2017vprop, khan2018fast}. 
Furthermore, curvature information can significantly improve learning dynamics in deep neural networks \citep{lin2023structured, lin2024remove, eschenhagen2024kroneckerfactored}, which has spawned recent work on software packages for scalable Hessian computations \citep{daxberger2022laplace, immer2021scalable}.
We make use of this development in our work to scale up SVN to BNN posteriors.

\section{Conclusion}
\label{sec:conclusion}

In this paper, we have introduced the SVN algorithm for BNN inference, which enhances traditional first-order methods by incorporating second-order curvature information through Hessian approximations. Our extensive evaluations across synthetic and real-world datasets, including UCI benchmarks and computer vision tasks, demonstrate that SVN consistently matches or outperforms existing particle-based methods like deep ensembles, repulsive ensembles, and SVGD, in terms of predictive performance, uncertainty estimation, and convergence speed. Additionally, our novel Last-Layer SVN variant proves computationally efficient for larger neural networks, maintaining competitive performance. These findings position SVN as a promising advancement in particle-based variational inference, offering more accurate and reliable deep learning models. \\

\paragraph{Limitations.}

We have demonstrated the competitive or superior performance of our SVN algorithm on a range of benchmark tasks, focusing on moderately-sized neural networks. While the Hessian approximations discussed in this work can be scaled to more complex models like GNNs, ViTs \citep{eschenhagen2024kroneckerfactored}, or LLMs \citep{yang2024bayesian, daxberger2022laplace, onal2024gaussianstochasticweightaveraging}, it remains unclear how well our method will perform in these scenarios. However, given the recent success of second-order methods in large NN training \citep{lin2023structured}, we remain optimistic. Additionally, while some Hessian approximations can be numerically unstable, we can mitigate this by selecting the appropriate approximation for each specific task.

\section*{Acknowledgments}
We want to thank Alexander Immer, Richard Paul, Emma Caldwell, Andrea Rubbi, and Arsen Sheverdin for helpful discussions. Moreover, we want to thank the Jülich Supercomputing Centre's support team, and in particular Rene Halver. VF was supported by a Branco Weiss Fellowship. This work was supported by Helmholtz AI computing resources (HAICORE) of the Helmholtz Association’s Initiative and Networking Fund through Helmholtz AI. This work was also supported by the Helmholtz Association Initiative and Networking Fund on the HAICORE@KIT partition.

\bibliographystyle{plainnat}

\begin{thebibliography}{87}
\providecommand{\natexlab}[1]{#1}
\providecommand{\url}[1]{\texttt{#1}}
\expandafter\ifx\csname urlstyle\endcsname\relax
  \providecommand{\doi}[1]{doi: #1}\else
  \providecommand{\doi}{doi: \begingroup \urlstyle{rm}\Url}\fi

\bibitem[Amari(1998)]{amari1998natural}
Shun‐ichi Amari.
\newblock Natural {G}radient {W}orks {E}fficiently in {L}earning.
\newblock \emph{Neural Computation}, 10:\penalty0 251--276, 1998.

\bibitem[Andrieu et~al.(2003)Andrieu, {De Freitas}, Doucet, and Jordan]{andrieu2003introduction}
Christophe Andrieu, Nando {De Freitas}, Arnaud Doucet, and {Michael I.} Jordan.
\newblock An {I}ntroduction to {MCMC} for {M}achine {L}earning.
\newblock \emph{Machine Learning}, 50\penalty0 (1-2):\penalty0 5--43, January 2003.
\newblock ISSN 1573-0565.
\newblock \doi{10.1023/A:1020281327116}.

\bibitem[Blei et~al.(2017)Blei, Kucukelbir, and McAuliffe]{blei2017variational}
David~M Blei, Alp Kucukelbir, and Jon~D McAuliffe.
\newblock Variational {I}nference: A {R}eview for {S}tatisticians.
\newblock \emph{Journal of the American statistical Association}, 112\penalty0 (518):\penalty0 859--877, 2017.

\bibitem[Botev et~al.(2017)Botev, Ritter, and Barber]{botev2017practical}
Aleksandar Botev, Hippolyt Ritter, and David Barber.
\newblock Practical {G}auss-{N}ewton {O}ptimisation for {D}eep {L}earning.
\newblock In Doina Precup and Yee~Whye Teh, editors, \emph{Proceedings of the 34th International Conference on Machine Learning}, volume~70 of \emph{Proceedings of Machine Learning Research}, pages 557--565. PMLR, 06--11 Aug 2017.

\bibitem[Brier(1950)]{brier_Score}
Glenn~W. Brier.
\newblock Verification of {F}orecasts {E}xpressed in {T}erms of {P}robability.
\newblock \emph{Monthly Weather Review}, 78\penalty0 (1):\penalty0 1 -- 3, 1950.

\bibitem[Chen and Ghattas(2020)]{chen2020projected_svgd}
Peng Chen and Omar Ghattas.
\newblock Projected {S}tein {V}ariational {G}radient {D}escent.
\newblock \emph{arXiv preprint arXiv:2002.03469}, 2020.

\bibitem[Chen et~al.(2020)Chen, Wu, Chen, O'Leary-Roseberry, and Ghattas]{chen2020projected_svn}
Peng Chen, Keyi Wu, Joshua Chen, Thomas O'Leary-Roseberry, and Omar Ghattas.
\newblock Projected {S}tein {V}ariational {N}ewton: {A} {F}ast and {S}calable {B}ayesian {I}nference {M}ethod in {H}igh {D}imensions, 2020.

\bibitem[Dangel et~al.(2020)Dangel, Kunstner, and Hennig]{dangel2020backpack}
Felix Dangel, Frederik Kunstner, and Philipp Hennig.
\newblock Backpack: Packing more into backprop.
\newblock In \emph{Proceedings of the 8th International Conference on Learning Representations (ICLR)}, 2020.

\bibitem[D'Angelo and Fortuin(2021{\natexlab{a}})]{dangelo2021annealed}
Francesco D'Angelo and Vincent Fortuin.
\newblock Annealed stein variational gradient descent.
\newblock \emph{arXiv preprint arXiv:2101.09815}, 2021{\natexlab{a}}.

\bibitem[D'Angelo and Fortuin(2021{\natexlab{b}})]{dangelo2023repulsive}
Francesco D'Angelo and Vincent Fortuin.
\newblock Repulsive {D}eep {E}nsembles are {B}ayesian.
\newblock In \emph{Neural Information Processing Systems}, 2021{\natexlab{b}}.

\bibitem[D'Angelo et~al.(2021)D'Angelo, Fortuin, and Wenzel]{d2021stein}
Francesco D'Angelo, Vincent Fortuin, and Florian Wenzel.
\newblock On {S}tein {V}ariational {N}eural {N}etwork {E}nsembles.
\newblock \emph{arXiv preprint arXiv:2106.10760}, 2021.

\bibitem[Daxberger et~al.(2021{\natexlab{a}})Daxberger, Kristiadi, Immer, Eschenhagen, Bauer, and Hennig]{daxberger2022laplace}
Erik Daxberger, Agustinus Kristiadi, Alexander Immer, Runa Eschenhagen, Matthias Bauer, and Philipp Hennig.
\newblock Laplace {R}edux - {E}ffortless {B}ayesian {D}eep {L}earning.
\newblock In M.~Ranzato, A.~Beygelzimer, Y.~Dauphin, P.S. Liang, and J.~Wortman Vaughan, editors, \emph{Advances in Neural Information Processing Systems}, volume~34, pages 20089--20103. Curran Associates, Inc., 2021{\natexlab{a}}.

\bibitem[Daxberger et~al.(2021{\natexlab{b}})Daxberger, Nalisnick, Allingham, Antorán, and Hernández-Lobato]{daxberger2021bayesian}
Erik Daxberger, Eric Nalisnick, James~Urquhart Allingham, Javier Antorán, and José~Miguel Hernández-Lobato.
\newblock Bayesian {D}eep {L}earning via {S}ubnetwork {I}nference.
\newblock In \emph{Proceedings of the 38th International Conference on Machine Learning (ICML)}, pages 2510--2521. PMLR, 2021{\natexlab{b}}.

\bibitem[Denker and LeCun(1990)]{denker1990transforming}
John Denker and Yann LeCun.
\newblock Transforming {N}eural-{N}et {O}utput {L}evels to {P}robability {D}istributions.
\newblock In R.P. Lippmann, J.~Moody, and D.~Touretzky, editors, \emph{Advances in Neural Information Processing Systems}, volume~3. Morgan-Kaufmann, 1990.

\bibitem[Detommaso et~al.(2018)Detommaso, Cui, Marzouk, Spantini, and Scheichl]{detommaso2018stein}
Gianluca Detommaso, Tiangang Cui, Youssef Marzouk, Alessio Spantini, and Robert Scheichl.
\newblock A {S}tein variational {N}ewton method.
\newblock In S.~Bengio, H.~Wallach, H.~Larochelle, K.~Grauman, N.~Cesa-Bianchi, and R.~Garnett, editors, \emph{Advances in Neural Information Processing Systems}, pages 9169--9179. Curran Associates, Inc., 2018.

\bibitem[Dua and Graff(2017)]{dua2017uci}
Dheeru Dua and Casey Graff.
\newblock {UCI} {M}achine {L}earning {R}epository, 2017.

\bibitem[Eschenhagen et~al.(2024)Eschenhagen, Immer, Turner, Schneider, and Hennig]{eschenhagen2024kroneckerfactored}
Runa Eschenhagen, Alexander Immer, Richard~E. Turner, Frank Schneider, and Philipp Hennig.
\newblock Kronecker-{F}actored {A}pproximate {C}urvature for {M}odern {N}eural {N}etwork {A}rchitectures, 2024.

\bibitem[Farquhar et~al.(2021)Farquhar, Smith, and Gal]{farquhar2021liberty}
Sebastian Farquhar, Lewis Smith, and Yarin Gal.
\newblock Liberty or {D}epth: {D}eep {B}ayesian {N}eural {N}ets {D}o {N}ot {N}eed {C}omplex {W}eight {P}osterior approximations, 2021.

\bibitem[Foong et~al.(2020)Foong, Burt, Li, and Turner]{foong2020expressiveness}
Andrew Y.~K. Foong, David~R. Burt, Yingzhen Li, and Richard~E. Turner.
\newblock On the {E}xpressiveness of {A}pproximate {I}nference in {B}ayesian {N}eural {N}etworks, 2020.

\bibitem[Gallego and Insua(2020)]{gallego2020stochastic}
Victor Gallego and David~Rios Insua.
\newblock Stochastic {G}radient {MCMC} with {R}epulsive {F}orces, 2020.

\bibitem[Garipov et~al.(2018)Garipov, Izmailov, Podoprikhin, Vetrov, and Wilson]{garipov2018loss}
Timur Garipov, Pavel Izmailov, Dmitrii Podoprikhin, Dmitry Vetrov, and Andrew~Gordon Wilson.
\newblock Loss {S}urfaces, {M}ode {C}onnectivity, and {F}ast {E}nsembling of {DNN}s.
\newblock \emph{arXiv preprint arXiv:1802.10026}, 2018.

\bibitem[Garriga-Alonso and Fortuin(2021)]{garriga2021exact}
Adri{\`a} Garriga-Alonso and Vincent Fortuin.
\newblock Exact langevin dynamics with stochastic gradients.
\newblock \emph{arXiv preprint arXiv:2102.01691}, 2021.

\bibitem[George et~al.(2018)George, Laurent, Bouthillier, Ballas, and Vincent]{kfac_eigendecomp}
Thomas George, C\'{e}sar Laurent, Xavier Bouthillier, Nicolas Ballas, and Pascal Vincent.
\newblock Fast {A}pproximate {N}atural {G}radient {D}escent in a {K}ronecker {F}actored {E}igenbasis.
\newblock In S.~Bengio, H.~Wallach, H.~Larochelle, K.~Grauman, N.~Cesa-Bianchi, and R.~Garnett, editors, \emph{Advances in Neural Information Processing Systems}, volume~31. Curran Associates, Inc., 2018.

\bibitem[Givens and Hoeting(2012)]{givens2012computational}
G.H. Givens and J.A. Hoeting.
\newblock \emph{Computational Statistics}.
\newblock Wiley Series in Computational Statistics. Wiley, 2012.
\newblock ISBN 9780470533314.

\bibitem[Goodfellow et~al.(2016)Goodfellow, Bengio, and Courville]{Goodfellow-et-al-2016}
Ian Goodfellow, Yoshua Bengio, and Aaron Courville.
\newblock \emph{Deep Learning}.
\newblock MIT Press, 2016.

\bibitem[Gruber and Buettner(2024)]{gruber2024better}
Sebastian~G. Gruber and Florian Buettner.
\newblock Better {U}ncertainty {C}alibration via {P}roper {S}cores for {C}lassification and {B}eyond, 2024.

\bibitem[Guo et~al.(2017)Guo, Pleiss, Sun, and Weinberger]{guo2017calibration}
Chuan Guo, Geoff Pleiss, Yu~Sun, and Kilian~Q. Weinberger.
\newblock On {C}alibration of {M}odern {N}eural {N}etworks.
\newblock In Doina Precup and Yee~Whye Teh, editors, \emph{Proceedings of the 34th International Conference on Machine Learning}, volume~70 of \emph{Proceedings of Machine Learning Research}, pages 1321--1330. PMLR, 06--11 Aug 2017.

\bibitem[Hansen and Salamon(1990)]{hansen90}
L.~K. Hansen and P.~Salamon.
\newblock Neural {N}etwork {E}nsembles.
\newblock \emph{IEEE Trans. Pattern Anal. Mach. Intell.}, 12\penalty0 (10):\penalty0 993–1001, October 1990.
\newblock ISSN 0162-8828.
\newblock \doi{10.1109/34.58871}.

\bibitem[Hempel et~al.(2023)Hempel, Sadeghi, and Kirsten]{app13126930}
Lars Hempel, Sina Sadeghi, and Toralf Kirsten.
\newblock Prediction of intensive care unit length of stay in the mimic-iv dataset.
\newblock \emph{Applied Sciences}, 13\penalty0 (12), 2023.
\newblock ISSN 2076-3417.
\newblock \doi{10.3390/app13126930}.
\newblock URL \url{https://www.mdpi.com/2076-3417/13/12/6930}.

\bibitem[Hernández-Lobato and Adams(2015)]{hernándezlobato2015probabilistic}
José~Miguel Hernández-Lobato and Ryan~P. Adams.
\newblock Probabilistic {B}ackpropagation for {S}calable {L}earning of {B}ayesian {N}eural {N}etworks, 2015.

\bibitem[Heskes(2000)]{heskes2000natural}
Tom Heskes.
\newblock On “{N}atural” {L}earning and {P}runing in {M}ultilayered {P}erceptrons.
\newblock \emph{Neural Computation}, 12\penalty0 (4):\penalty0 881--901, 2000.

\bibitem[Huang et~al.(2023)Huang, Chouzenoux, Elvira, and Pesquet]{huang2023efficient}
Yunshi Huang, Emilie Chouzenoux, Victor Elvira, and Jean-Christophe Pesquet.
\newblock Efficient {B}ayes {I}nference in {N}eural {N}etworks through {A}daptive {I}mportance {S}ampling, 2023.

\bibitem[Immer et~al.(2021{\natexlab{a}})Immer, Bauer, Fortuin, R{\"a}tsch, and Khan]{immer2021scalable}
Alexander Immer, Matthias Bauer, Vincent Fortuin, Gunnar R{\"a}tsch, and Mohammad~Emtiyaz Khan.
\newblock Scalable {M}arginal {L}ikelihood {E}stimation for {M}odel {S}election in {D}eep {L}earning.
\newblock \emph{arXiv preprint arXiv:2104.04975}, 2021{\natexlab{a}}.

\bibitem[Immer et~al.(2021{\natexlab{b}})Immer, Korzepa, and Bauer]{immer2021improving}
Alexander Immer, Maciej Korzepa, and Matthias Bauer.
\newblock Improving predictions of {B}ayesian neural nets via local linearization.
\newblock In \emph{International Conference on Artificial Intelligence and Statistics}, pages 703--711. PMLR, 2021{\natexlab{b}}.

\bibitem[Izmailov et~al.(2021)Izmailov, Vikram, Hoffman, and Wilson]{izmailov2021bayesian}
Pavel Izmailov, Sharad Vikram, Matthew~D Hoffman, and Andrew Gordon~Gordon Wilson.
\newblock What {A}re {B}ayesian {N}eural {N}etwork {P}osteriors {R}eally {L}ike?
\newblock In Marina Meila and Tong Zhang, editors, \emph{Proceedings of the 38th International Conference on Machine Learning}, volume 139 of \emph{Proceedings of Machine Learning Research}, pages 4629--4640. PMLR, 18--24 Jul 2021.

\bibitem[Jain(1989)]{jain1989fundamentals}
A.K. Jain.
\newblock \emph{Fundamentals of Digital Image Processing}.
\newblock Prentice-Hall information and system sciences series. Prentice Hall, 1989.
\newblock ISBN 9780133361650.

\bibitem[Johnson et~al.(2024)Johnson, Bulgarelli, Pollard, Gow, Moody, Horng, Celi, and Mark]{johnson2024mimic}
Alistair Johnson, Lucas Bulgarelli, Tom Pollard, Brian Gow, Benjamin Moody, Steven Horng, Leo~Anthony Celi, and Roger Mark.
\newblock Mimic-iv, 2024.
\newblock URL \url{https://doi.org/10.13026/hxp0-hg59}.

\bibitem[Khan et~al.(2018)Khan, Nielsen, Tangkaratt, Lin, Gal, and Srivastava]{khan2018fast}
Mohammad Khan, Didrik Nielsen, Voot Tangkaratt, Wu~Lin, Yarin Gal, and Akash Srivastava.
\newblock Fast and scalable {B}ayesian {D}eep {L}earning by {W}eight-{P}erturbation in {A}dam.
\newblock In Jennifer Dy and Andreas Krause, editors, \emph{Proceedings of the 35th International Conference on Machine Learning}, volume~80 of \emph{Proceedings of Machine Learning Research}, pages 2611--2620. PMLR, 10--15 Jul 2018.

\bibitem[Khan et~al.(2017{\natexlab{a}})Khan, Lin, Tangkaratt, Liu, and Nielsen]{khan2017variational}
Mohammad~Emtiyaz Khan, Wu~Lin, Voot Tangkaratt, Zuozhu Liu, and Didrik Nielsen.
\newblock Variational {A}daptive-{N}ewton {M}ethod for {E}xplorative {L}earning, 2017{\natexlab{a}}.

\bibitem[Khan et~al.(2017{\natexlab{b}})Khan, Liu, Tangkaratt, and Gal]{khan2017vprop}
Mohammad~Emtiyaz Khan, Zuozhu Liu, Voot Tangkaratt, and Yarin Gal.
\newblock Vprop: {V}ariational {I}nference using {RMS}prop, 2017{\natexlab{b}}.

\bibitem[Kobayashi et~al.(2022)Kobayashi, Aceituno, and von Oswald]{kobayashi2022disentangling}
Seijin Kobayashi, Pau~Vilimelis Aceituno, and Johannes von Oswald.
\newblock Disentangling the {P}redictive {V}ariance of {D}eep {E}nsembles through the {N}eural {T}angent {K}ernel, 2022.

\bibitem[Kristiadi et~al.(2020)Kristiadi, Hein, and Hennig]{kristiadi2020being}
Agustinus Kristiadi, Matthias Hein, and Philipp Hennig.
\newblock Being {B}ayesian, {E}ven {J}ust a {B}it, {F}ixes {O}verconfidence in {R}e{LU} networks.
\newblock In Hal~Daumé III and Aarti Singh, editors, \emph{Proceedings of the 37th International Conference on Machine Learning}, volume 119 of \emph{Proceedings of Machine Learning Research}, pages 5436--5446. PMLR, 13--18 Jul 2020.

\bibitem[Kullback and Leibler(1951)]{kl_divergence}
S.~Kullback and R.~A. Leibler.
\newblock {On Information and Sufficiency}.
\newblock \emph{The Annals of Mathematical Statistics}, 22\penalty0 (1):\penalty0 79 -- 86, 1951.

\bibitem[Lakshminarayanan et~al.(2017)Lakshminarayanan, Pritzel, and Blundell]{lakshminarayanan2017simple}
Balaji Lakshminarayanan, Alexander Pritzel, and Charles Blundell.
\newblock Simple and {S}calable {P}redictive {U}ncertainty {E}stimation using {D}eep {E}nsembles.
\newblock In \emph{Advances in neural information processing systems}, pages 6402--6413, 2017.

\bibitem[Laplace(1774)]{laplace1774memoires}
Pierre-Simon Laplace.
\newblock \emph{M{\'e}moires de Math{\'e}matique et de Physique, Tome Sixieme}.
\newblock Imprimerie Royale, 1774.

\bibitem[LeCun(1998)]{lecun1998mnist}
Yann LeCun.
\newblock The {MNIST} {D}atabase of {H}andwritten {D}igits.
\newblock 1998.

\bibitem[LeCun et~al.(1989)LeCun, Denker, and Solla]{lecun1990optimal}
Yann LeCun, John Denker, and Sara Solla.
\newblock Optimal {B}rain {D}amage.
\newblock In D.~Touretzky, editor, \emph{Advances in Neural Information Processing Systems}, volume~2. Morgan-Kaufmann, 1989.

\bibitem[LeCun et~al.(1998)LeCun, Bottou, Bengio, and Haffner]{lenet}
Yann LeCun, L{\'e}on Bottou, Yoshua Bengio, and Patrick Haffner.
\newblock Gradient-{B}ased {L}earning {A}pplied to {D}ocument {R}ecognition.
\newblock \emph{Proceedings of the IEEE}, 86\penalty0 (11):\penalty0 2278--2323, 1998.
\newblock ISSN 0018-9219.
\newblock \doi{10.1109/5.726791}.

\bibitem[Lee et~al.(2020)Lee, Humt, Feng, and Triebel]{lee2020estimating}
Jongseok Lee, Matthias Humt, Jianxiang Feng, and Rudolph Triebel.
\newblock Estimating {M}odel {U}ncertainty of {N}eural {N}etworks in {S}parse {I}nformation {F}orm.
\newblock In Hal~Daumé III and Aarti Singh, editors, \emph{Proceedings of the 37th International Conference on Machine Learning}, volume 119 of \emph{Proceedings of Machine Learning Research}, pages 5702--5713. PMLR, 13--18 Jul 2020.

\bibitem[Leviyev et~al.(2022)Leviyev, Chen, Wang, Ghattas, and Zimmerman]{leviyev2022stochastic}
Alex Leviyev, Joshua Chen, Yifei Wang, Omar Ghattas, and Aaron Zimmerman.
\newblock A stochastic {S}tein {V}ariational {N}ewton method, 2022.

\bibitem[Li et~al.(2018)Li, Xu, Taylor, Studer, and Goldstein]{li2018visualizing}
Hao Li, Zheng Xu, Gavin Taylor, Christoph Studer, and Tom Goldstein.
\newblock Visualizing the {L}oss {L}andscape of {N}eural {N}ets, 2018.

\bibitem[Lin et~al.(2023)Lin, Dangel, Eschenhagen, Neklyudov, Kristiadi, Turner, and Makhzani]{lin2023structured}
Wu~Lin, Felix Dangel, Runa Eschenhagen, Kirill Neklyudov, Agustinus Kristiadi, Richard~E. Turner, and Alireza Makhzani.
\newblock Structured {I}nverse-{F}ree {N}atural {G}radient: {M}emory-{E}fficient \& {N}umerically-{S}table {KFAC} for {L}arge {N}eural {N}ets, 2023.

\bibitem[Lin et~al.(2024)Lin, Dangel, Eschenhagen, Bae, Turner, and Makhzani]{lin2024remove}
Wu~Lin, Felix Dangel, Runa Eschenhagen, Juhan Bae, Richard~E. Turner, and Alireza Makhzani.
\newblock Can {W}e {R}emove the {S}quare-{R}oot in {A}daptive {G}radient {M}ethods? {A} {S}econd-{O}rder {P}erspective, 2024.

\bibitem[Liu and Zhu(2017)]{liu2017riemannian}
Chang Liu and Jun Zhu.
\newblock Riemannian {S}tein {V}ariational {G}radient {D}escent for {B}ayesian {I}nference, 2017.

\bibitem[Liu et~al.(2019)Liu, Zhuo, Cheng, Zhang, and Zhu]{liu2019understanding}
Chang Liu, Jingwei Zhuo, Pengyu Cheng, Ruiyi Zhang, and Jun Zhu.
\newblock Understanding and {A}ccelerating {P}article-{B}ased {V}ariational {I}nference.
\newblock In Kamalika Chaudhuri and Ruslan Salakhutdinov, editors, \emph{Proceedings of the 36th International Conference on Machine Learning}, volume~97 of \emph{Proceedings of Machine Learning Research}, pages 4082--4092. PMLR, 09--15 Jun 2019.

\bibitem[Liu and Wang(2016)]{liu2016stein}
Qiang Liu and Dilin Wang.
\newblock Stein {V}ariational {G}radient {D}escent: A {G}eneral {P}urpose {B}ayesian {I}nference {A}lgorithm.
\newblock In \emph{Advances in neural information processing systems}, pages 2378--2386, 2016.

\bibitem[MacKay(1992{\natexlab{a}})]{mackay1992bayesian}
David~JC MacKay.
\newblock Bayesian {I}nterpolation.
\newblock \emph{Neural Computation}, 4\penalty0 (3):\penalty0 415--447, 1992{\natexlab{a}}.

\bibitem[MacKay(1992{\natexlab{b}})]{mackay1992practical}
David~JC MacKay.
\newblock A {P}ractical {B}ayesian {F}ramework for {B}ackpropagation {N}etworks.
\newblock \emph{Neural computation}, 4\penalty0 (3):\penalty0 448--472, 1992{\natexlab{b}}.

\bibitem[Martens(2020)]{martens2020new}
James Martens.
\newblock New insights and perspectives on the natural gradient method.
\newblock \emph{Journal of Machine Learning Research}, 21\penalty0 (146):\penalty0 1--76, 2020.

\bibitem[Martens and Grosse(2015)]{martens2015optimizing}
James Martens and Roger Grosse.
\newblock Optimizing {N}eural {N}etworks with {K}ronecker-{F}actored {A}pproximate {C}urvature.
\newblock In \emph{Proceedings of the 32nd International Conference on Machine Learning (ICML)}, pages 2408--2417. PMLR, 2015.

\bibitem[Neal(1995)]{neal1995bayesian}
Radford~M. Neal.
\newblock Bayesian {L}earning for {N}eural {N}etworks.
\newblock 1995.

\bibitem[Nix and Weigend(1994)]{gaussian_nll_loss}
D.A. Nix and A.S. Weigend.
\newblock Estimating the mean and variance of the target probability distribution.
\newblock In \emph{Proceedings of 1994 IEEE International Conference on Neural Networks (ICNN'94)}, volume~1, pages 55--60 vol.1, 1994.
\newblock \doi{10.1109/ICNN.1994.374138}.

\bibitem[Njieutcheu~Tassi et~al.(2022)Njieutcheu~Tassi, Gawlikowski, Fitri, and Triebel]{dlr187799}
Cedrique~Rovile Njieutcheu~Tassi, Jakob Gawlikowski, Auliya~Unnisa Fitri, and Rudolph Triebel.
\newblock The impact of averaging logits over probabilities on ensembles of neural networks.
\newblock In \emph{2022 Workshop on Artificial Intelligence Safety, AISafety 2022}, CEUR Workshop Proceedings, 2022.

\bibitem[Ober and Rasmussen(2019)]{ober2019benchmarking}
Sebastian~W. Ober and Carl~Edward Rasmussen.
\newblock Benchmarking the {N}eural {L}inear {M}odel for {R}egression, 2019.

\bibitem[Onal et~al.(2024)Onal, Flöge, Caldwell, Sheverdin, and Fortuin]{onal2024gaussianstochasticweightaveraging}
Emre Onal, Klemens Flöge, Emma Caldwell, Arsen Sheverdin, and Vincent Fortuin.
\newblock Gaussian {S}tochastic {W}eight {A}veraging for {B}ayesian {L}ow-{R}ank {A}daptation of {L}arge {L}anguage {M}odels, 2024.

\bibitem[Osawa et~al.(2023)Osawa, Ishikawa, Yokota, Li, and Hoefler]{osawa2023asdl}
Kazuki Osawa, Satoki Ishikawa, Rio Yokota, Shigang Li, and Torsten Hoefler.
\newblock {ASDL}: A {U}nified {I}nterface for {G}radient {P}reconditioning in {P}y{T}orch, 2023.

\bibitem[Papamarkou et~al.(2021)Papamarkou, Hinkle, Young, and Womble]{papamarkou2021challenges}
Theodore Papamarkou, Jacob Hinkle, M.~Todd Young, and David Womble.
\newblock Challenges in {M}arkov chain {M}onte {C}arlo for {B}ayesian neural networks, 2021.

\bibitem[Papamarkou et~al.(2024)Papamarkou, Skoularidou, Palla, Aitchison, Arbel, Dunson, Filippone, Fortuin, Hennig, Hubin, Immer, Karaletsos, Khan, Kristiadi, Li, Mandt, Nemeth, Osborne, Rudner, Rügamer, Teh, Welling, Wilson, and Zhang]{papamarkou2024position}
Theodore Papamarkou, Maria Skoularidou, Konstantina Palla, Laurence Aitchison, Julyan Arbel, David Dunson, Maurizio Filippone, Vincent Fortuin, Philipp Hennig, Aliaksandr Hubin, Alexander Immer, Theofanis Karaletsos, Mohammad~Emtiyaz Khan, Agustinus Kristiadi, Yingzhen Li, Stephan Mandt, Christopher Nemeth, Michael~A Osborne, Tim~GJ Rudner, David Rügamer, Yee~Whye Teh, Max Welling, Andrew~Gordon Wilson, and Ruqi Zhang.
\newblock Position {P}aper: {B}ayesian {D}eep {L}earning in the {A}ge of {L}arge-{S}cale {AI}.
\newblock \emph{arXiv preprint arXiv:2402.00809}, 2024.

\bibitem[Paszke et~al.(2019)Paszke, Gross, Massa, Lerer, Bradbury, Chanan, Killeen, Lin, Gimelshein, Antiga, Desmaison, Köpf, Yang, DeVito, Raison, Tejani, Chilamkurthy, Steiner, Fang, Bai, and Chintala]{paszke2019pytorch}
Adam Paszke, Sam Gross, Francisco Massa, Adam Lerer, James Bradbury, Gregory Chanan, Trevor Killeen, Zeming Lin, Natalia Gimelshein, Luca Antiga, Alban Desmaison, Andreas Köpf, Edward Yang, Zach DeVito, Martin Raison, Alykhan Tejani, Sasank Chilamkurthy, Benoit Steiner, Lu~Fang, Junjie Bai, and Soumith Chintala.
\newblock {P}y{T}orch: {A}n {I}mperative {S}tyle, {H}igh-{P}erformance {D}eep {L}earning {L}ibrary, 2019.

\bibitem[Pearson(1895)]{Pearson1895}
Karl Pearson.
\newblock Contributions to the {M}athematical {T}heory of {E}volution, {II}: {S}kew {V}ariation in {H}omogeneous {M}aterial.
\newblock \emph{Philosophical Transactions of the Royal Society}, 186:\penalty0 343--414, 1895.
\newblock \doi{10.1098/rsta.1895.0010}.

\bibitem[Pedregosa et~al.(2011)Pedregosa, Varoquaux, Gramfort, Michel, Thirion, Grisel, Blondel, Prettenhofer, Weiss, Dubourg, Vanderplas, Passos, Cournapeau, Brucher, Perrot, and Duchesnay]{scikit-learn}
F.~Pedregosa, G.~Varoquaux, A.~Gramfort, V.~Michel, B.~Thirion, O.~Grisel, M.~Blondel, P.~Prettenhofer, R.~Weiss, V.~Dubourg, J.~Vanderplas, A.~Passos, D.~Cournapeau, M.~Brucher, M.~Perrot, and E.~Duchesnay.
\newblock Scikit-learn: Machine {L}earning in {P}ython.
\newblock \emph{Journal of Machine Learning Research}, 12:\penalty0 2825--2830, 2011.

\bibitem[Pop and Fulop(2018)]{pop2018deep}
Remus Pop and Patric Fulop.
\newblock Deep {E}nsemble {B}ayesian {A}ctive {L}earning : {A}ddressing the {M}ode {C}ollapse issue in {M}onte {C}arlo dropout via {E}nsembles, 2018.

\bibitem[Ritter et~al.(2018{\natexlab{a}})Ritter, Botev, and Barber]{ritter2018online}
Hippolyt Ritter, Aleksandar Botev, and David Barber.
\newblock Online {S}tructured {L}aplace {A}pproximations {F}or {O}vercoming {C}atastrophic {F}orgetting.
\newblock In \emph{Advances in Neural Information Processing Systems 31 (NeurIPS)}, 2018{\natexlab{a}}.

\bibitem[Ritter et~al.(2018{\natexlab{b}})Ritter, Botev, and Barber]{ritter2018scalable}
Hippolyt Ritter, Aleksandar Botev, and David Barber.
\newblock A {S}calable {L}aplace {A}pproximation for {N}eural {N}etworks.
\newblock In \emph{Proceedings of the 6th International Conference on Learning Representations (ICLR)}, 2018{\natexlab{b}}.

\bibitem[Rogozhnikov(2022)]{rogozhnikov2022einops}
Alex Rogozhnikov.
\newblock Einops: {C}lear and {R}eliable {T}ensor {M}anipulations with {E}instein-like {N}otation.
\newblock In \emph{International Conference on Learning Representations}, 2022.

\bibitem[Schraudolph(2002)]{Schraudolph_ggn}
Nicol Schraudolph.
\newblock Fast {C}urvature {M}atrix-{V}ector {P}roducts for {S}econd-{O}rder {G}radient {D}escent.
\newblock \emph{Neural computation}, 14:\penalty0 1723--38, 08 2002.

\bibitem[Silverman(1986)]{Silverman86}
B.~W. Silverman.
\newblock \emph{Density Estimation for Statistics and Data Analysis}.
\newblock Chapman \& Hall, London, 1986.

\bibitem[Virtanen et~al.(2020)Virtanen, Gommers, Oliphant, Haberland, Reddy, Cournapeau, Burovski, Peterson, Weckesser, Bright, {van der Walt}, Brett, Wilson, Millman, Mayorov, Nelson, Jones, Kern, Larson, Carey, Polat, Feng, Moore, {VanderPlas}, Laxalde, Perktold, Cimrman, Henriksen, Quintero, Harris, Archibald, Ribeiro, Pedregosa, {van Mulbregt}, and {SciPy 1.0 Contributors}]{2020SciPy-NMeth}
Pauli Virtanen, Ralf Gommers, Travis~E. Oliphant, Matt Haberland, Tyler Reddy, David Cournapeau, Evgeni Burovski, Pearu Peterson, Warren Weckesser, Jonathan Bright, St{\'e}fan~J. {van der Walt}, Matthew Brett, Joshua Wilson, K.~Jarrod Millman, Nikolay Mayorov, Andrew R.~J. Nelson, Eric Jones, Robert Kern, Eric Larson, C~J Carey, {\.I}lhan Polat, Yu~Feng, Eric~W. Moore, Jake {VanderPlas}, Denis Laxalde, Josef Perktold, Robert Cimrman, Ian Henriksen, E.~A. Quintero, Charles~R. Harris, Anne~M. Archibald, Ant{\^o}nio~H. Ribeiro, Fabian Pedregosa, Paul {van Mulbregt}, and {SciPy 1.0 Contributors}.
\newblock {{SciPy} 1.0: Fundamental Algorithms for Scientific Computing in Python}.
\newblock \emph{Nature Methods}, 17:\penalty0 261--272, 2020.

\bibitem[Welling and Teh(2011)]{welling2011bayesian}
Max Welling and Yee~W Teh.
\newblock Bayesian {L}earning via {S}tochastic {G}radient {L}angevin {D}ynamics.
\newblock In \emph{Proceedings of the 28th international conference on machine learning}, pages 681--688, 2011.

\bibitem[Wenzel et~al.(2020{\natexlab{a}})Wenzel, Roth, Veeling, {\'S}wiatkowski, Tran, Mandt, Snoek, Salimans, Jenatton, and Nowozin]{wenzel2020good}
Florian Wenzel, Kevin Roth, Bastiaan~S Veeling, Jakub {\'S}wiatkowski, Linh Tran, Stephan Mandt, Jasper Snoek, Tim Salimans, Rodolphe Jenatton, and Sebastian Nowozin.
\newblock How {G}ood is the {B}ayes {P}osterior in {D}eep {N}eural {N}etworks {R}eally?
\newblock \emph{arXiv preprint arXiv:2002.02405}, 2020{\natexlab{a}}.

\bibitem[Wenzel et~al.(2020{\natexlab{b}})Wenzel, Snoek, Tran, and Jenatton]{wenzel2020hyper}
Florian Wenzel, Jasper Snoek, Dustin Tran, and Rodolphe Jenatton.
\newblock Hyperparameter {E}nsembles for {R}obustness and {U}ncertainty {Q}uantification.
\newblock In \emph{Advances in Neural Information Processing Systems}, 2020{\natexlab{b}}.

\bibitem[Wild et~al.(2023)Wild, Ghalebikesabi, Sejdinovic, and Knoblauch]{wild2023a}
Veit~David Wild, Sahra Ghalebikesabi, Dino Sejdinovic, and Jeremias Knoblauch.
\newblock A {R}igorous {L}ink between {D}eep {E}nsembles and ({V}ariational) {B}ayesian {M}ethods.
\newblock In \emph{Thirty-seventh Conference on Neural Information Processing Systems}, 2023.

\bibitem[Winter et~al.(2023)Winter, Hartwig, and Kirsten]{winter2023predicting}
Alexander Winter, Mattis Hartwig, and Toralf Kirsten.
\newblock Predicting hospital length of stay of patients leaving the emergency department.
\newblock In \emph{Proceedings of the 16th International Joint Conference on Biomedical Engineering Systems and Technologies (BIOSTEC 2023) - Volume 5: HEALTHINF}, pages 124--131. SCITEPRESS, 2023.
\newblock ISBN 978-989-758-631-6.
\newblock \doi{10.5220/0011671700003414}.

\bibitem[Wright and Nocedal(1999)]{Wright1999NumericalOptimization}
Stephen Wright and Jorge Nocedal.
\newblock \emph{Numerical Optimization}.
\newblock Springer Science, 1999.

\bibitem[Xiao et~al.(2017)Xiao, Rasul, and Vollgraf]{xiao2017fashion}
Han Xiao, Kashif Rasul, and Roland Vollgraf.
\newblock Fashion-{MNIST}: a {N}ovel {I}mage {D}ataset for {B}enchmarking {M}achine {L}earning {A}lgorithms.
\newblock \emph{arXiv preprint arXiv:1708.07747}, 2017.

\bibitem[Yang et~al.(2024)Yang, Robeyns, Wang, and Aitchison]{yang2024bayesian}
Adam~X. Yang, Maxime Robeyns, Xi~Wang, and Laurence Aitchison.
\newblock Bayesian {L}ow-rank {A}daptation for {L}arge {L}anguage {M}odels, 2024.

\bibitem[Zhuo et~al.(2018)Zhuo, Liu, Shi, Zhu, Chen, and Zhang]{zhuo2018message}
Jingwei Zhuo, Chang Liu, Jiaxin Shi, Jun Zhu, Ning Chen, and Bo~Zhang.
\newblock Message {P}assing {S}tein {V}ariational {G}radient {D}escent.
\newblock In \emph{International Conference on Machine Learning}, pages 6018--6027. PMLR, 2018.

\end{thebibliography}

\newpage
\appendix
\onecolumn

\counterwithin{figure}{section}
\counterwithin{table}{section}
\counterwithin{equation}{section}

\section{Hessian Approximation Details}
\label{app: hessian approx details}

\paragraph{Diagonal}

The Diagonal approximation disregards off-diagonal elements in the Fisher Information Matrix $ \mathbf{F}_\phi$. As such we can rewrite \cref{eq: fisher} as: 

\begin{equation}
\mathbf{F}_\phi = \,  \sum_{i=1}^{b} \mathbb{E}_{\tilde{y} \sim p(y | g_\phi(x_i))} \left[ \nabla_\phi (\log p(\tilde{y}|g_\phi(x_i)) ) \odot
\nabla_\phi (\log p(\tilde{y}|g_\phi(x_i)) ) \, \right],
\end{equation}

where $\odot$ represents the element-wise or Hadamard product. While it has been suggested to refrain from this approximation due to its oversimplification \citep{mackay1992practical}, the computational benefits are immense. Inverting a diagonal Fisher $ \mathbf{F}_\phi$, $\phi \in \mathbb{R}^d$, can be done in $\mathcal{O}(d)$, which is considerably faster than for the full Hessian, $\mathcal{O}(d^3)$. Moreover, recent works suggest that for sufficiently large neural networks, the approximation can yield good results \citep{farquhar2021liberty}.

\paragraph{KFAC}
The KFAC factorization serves as a balance between two extremes: diagonal factorization, which can be overly restrictive, and the full Fisher matrix, which is computationally impractical. The main idea is to capture the correlation between weights within the same layer while assuming that weights from different layers are independent. This assumption is more refined than the diagonal factorization, where all weights are considered independent. We are approximating the GGN $\mathbf{H}_\phi$ in the $l$-th layer with parameters $\phi_l$ as

\[
\mathbf{H}_{\phi_l} = \mathbf{Q}_{\phi_l} \otimes \mathbf{K}_{\phi_l}.
\]

Here, the factors are computed with the outer products of pre-activations and Jacobians with respect to the output of a layer \citep{martens2015optimizing, botev2017practical}. Following \citet{kfac_eigendecomp}, the Kronecker factors are represented in their eigendecompositions in \citet{daxberger2022laplace}.

\section{SVN Algorithm Details}
\label{app: SVN algo details}

We implemented all algorithms using the popular PyTorch \citep{paszke2019pytorch} framework. We also make extensive use of the EinOps \citep{rogozhnikov2022einops} library to handle some of the more involved tensor operations. The Hessian computations are done using the Laplace-Redux library \citep{daxberger2022laplace}, which itself uses the Backpack \citep{dangel2020backpack} and ASDL \citep{osawa2023asdl} libraries in its backend. In order to solve the linear systems presented in \cref{subsec: SVN}, we utilise solvers implemented in the SciPy \citep{2020SciPy-NMeth}. 

\paragraph{Block-Diagonal Approximation}

In the block-diagonal approximation, we are just considering the diagonal entries of the SVN Hessian, which uses the following block-diagonal SVN Hessian

\[
H^{\operatorname{SVN}}:=\left[\begin{array}{ccc}
h^{11} & \cdots & 0 \\
\vdots & \ddots & \vdots \\
0 & \cdots & h^{N N}
\end{array}\right]
\]

in \cref{eq: linear system}. This results in solving \cref{eq: block diag linear system} for each particle independently. This approach has significant computational advantages over the full SVN Hessian $H^{\operatorname{SVN}}$.

\paragraph{Linear System Solver}

In order to solve the full linear system in \cref{eq: linear system} or the block diagonal version \cref{eq: block diag linear system}, we utilize SciPy's Conjugate Newton method. The conjugate Newton method has strong theoretical grounding \citep{Wright1999NumericalOptimization} for sufficient numbers of iterations and was used by \citet{detommaso2018stein} in their original implementation as well.  We use $50$ iterations for all our computations. \citet{leviyev2022stochastic} adjusted the Hessian with a Levenberg-like damping term and then employed a Cholesky decomposition to solve the resulting linear system. However, we will leave this for future work. 

\paragraph{Matrix-vector products for different Hessian Approximations}

In order to use the numerical techniques discussed above to solve the linear systems in \cref{eq: linear system}, \cref{eq: block diag linear system}, one needs to define a Hessian-vector product for $\mathbf{H}_\phi \in \mathbb{R}^{d \times d}$ and $x \in \mathbb{R}^d$. For the full Hessian approximation, this is just the standard matrix-vector product. In the diagonal approximation, we just consider the diagonal elements, which means that $\mathbf{H}^{\text{Diag}}_\phi \in \mathbb{R}^d$ is a vector. As such, the matrix-vector product collapses to a simple Hadamard product of these two vectors: 

\begin{equation}
\mathbf{H}^{\text{Diag}}_\phi x = \mathbf{H}^{\text{Diag}}_\phi \odot x
\end{equation}

The KFAC approximation involves a layer-wise block diagonal approximation. For the parameter vector $\phi_l$ representing the $l$-th layer, we aim to calculate $\mathbf{H}_{\phi_l} \phi_l$. The Hessian of the $l$-th layer is decomposed into its Kronecker factors: $\mathbf{H}^{\text{KFAC}}_{\phi_l} = \mathbf{Q}_{\phi_l} \otimes \mathbf{K}_{\phi_l}$, which are represented in their eigendecomposition as $\mathbf{Q}_{\phi_l} = V_Q \Lambda_Q V_Q^\top$ and $\mathbf{K}_{\phi_l} = V_K \Lambda_K V_K^\top$. For notational convenience, we omit the dependency on $\phi_l$. 

A straightforward approach to $\mathbf{H}^{\text{KFAC}}_{\phi_l} x$ would involve using the eigenvectors $(V_Q, V_K)$ and eigenvalues $(\Lambda_Q, \Lambda_K)$ to construct $\mathbf{Q}_{\phi_l}$ and $\mathbf{K}_{\phi_l}$, then explicitly calculate the Kronecker product $\mathbf{Q}_{\phi_l} \otimes \mathbf{K}_{\phi_l}$ and perform a standard matrix-vector product. However, this approach is highly inefficient in PyTorch. Instead, we utilize the ``vec trick'' \citep{jain1989fundamentals} for Kronecker products. With $Q \in \mathbb{R}^a$ and $K \in \mathbb{R}^b$, we have $ab = |\phi_l| = |x|$ and write $X \in \mathbb{R}^{a \times b}$ as a reshaped version of $x$. The ``vec trick'' computes our desired product as follows:

\begin{align}
    v &= (Q \otimes K) x \\
    &= (Q \otimes K^\top) x \\
    &= \text{vec}((Q X K)^\top) \\
    &= \text{vec}((((Q^\top) X^\top) K^\top)).
\end{align}

This approach allows us to use the Kronecker factors exactly without explicitly constructing $\mathbf{H}^{\text{KFAC}}_{\phi_l}$ using the Kronecker product.

\paragraph{LL-SVN Algorithm}

\cref{alg: LL-SVN algorithm} details our novel last-layer SVN algorithm. Conceptually, it is very similar to the SVN Ensembles algorithm described in \cref{sec:svn algo}. The main difference is that it restricts the computation of the SVN update to only the parameters in the last layer, while using \( v^{\operatorname{SVGD}}_l \) for the rest of the parameters. Since our SVN algorithm also needs to compute \( v^{\operatorname{SVGD}}_l \) for the entire network, this part remains identical in both algorithms. The only difference is that we approximate the Hessian only for the parameters in the last layer, \( \phi^{\text{LL}} \), using the KFAC approximation. We then solve the full linear system in \cref{eq: linear system} or block diagonal linear systems in \cref{eq: block diag linear system} only with the \( \phi^{\text{LL}} \) parameters.

\begin{algorithm}[ht]
\caption{One iteration of our LL-SVN neural network ensemble algorithm}
\begin{algorithmic}[1]
\STATE \textbf{Input:} Particles $\{\phi_i^l\}_{i=1}^n$ at stage $l$; step size $\varepsilon$, Kernel $k(\cdot, \cdot) \in \{ k^{M_{\text{SVN}}}, k^I \}$, Hessian Approximation $\mathbf{H} \in \{\mathbf{H}_{\text{Full}}, \mathbf{H}_{\text{KFAC}}, \mathbf{H}_{\text{Diag}} \}$, number of parameters in last layer $d_{ll}$
\STATE \textbf{Output:} Particles $\{\phi_i^{l+1}\}_{i=1}^n$ at stage $l+1$
\STATE Compute $v^{\operatorname{SVGD}}_l$
\FOR{$i=1,2,\ldots, n$}
    \STATE Obtain last-layer weights: $\phi_i = \phi_i^{FL} \cup \phi_i^{LL}$
    \IF{Block Diagonal Approximation}  
        \STATE Solve linear system \cref{eq: block diag linear system} for $\alpha^1, \ldots, \alpha^n$ with $\phi_i^{LL}$
    \ELSE
        \STATE Solve linear system \cref{eq: linear system} for $\alpha^1, \ldots, \alpha^n$ with $\phi_i^{LL}$
    \ENDIF
    \STATE Concatenate: $v_l^{\operatorname{LL-SVN}}\left(\phi_i^{k} \right) = v^{\operatorname{SVGD}}(\phi_i^{FL}) \cup v_l^{\operatorname{SVN}}(\phi_i^{LL}) $
    \STATE Set $\phi_i^{k+1} \leftarrow \phi_i^k + \varepsilon \, v_l^{\operatorname{LL-SVN}}\left(\phi_i^{k} \right)$ given $\alpha^1, \ldots, \alpha^n$
\ENDFOR
\end{algorithmic}
\label{alg: LL-SVN algorithm}
\end{algorithm}

\section{Experiment Details}
\label{app: experiment details}

\subsection{Toy Regression Details}
\label{app sub sec: toy regression}

The dataset for this study is a 1D regression dataset generated using a specific mapping function and domain. We use \texttt{task\_set=4} of the Toy Regression presented in \citet{dangelo2023repulsive}. The dataset characteristics are as follows:

The mapping function used to generate the dataset is given by:
\begin{equation}
    f(x) = (x - 3)^3
\end{equation}

The dataset is created within the following domains:
\begin{itemize}
    \item \textbf{Training Domain:} \([2, 3] \cup [4.5, 6]\)
    \item \textbf{Test Domain:} \([0, 7]\)
\end{itemize}

To simulate real-world data variations, Gaussian noise with a standard deviation of 0.25 is added to the function values. The dataset includes:
\begin{itemize}
    \item \textbf{Number of Training Samples:} 150
    \item \textbf{Number of Test Samples:} 200
\end{itemize}

Additionally, specific clusters of data points, referred to as blobs, are included within the training domain at the intervals \([1.5, 2.5]\) and \([4.5, 6.0]\). These blobs introduce additional complexity to the dataset, providing a more challenging environment for evaluating regression models.

The dataset generation process is controlled by a fixed random seed of 42 to ensure reproducibility. The training, testing, and validation sets are drawn from their respective domains, incorporating noise and blobs to enhance the dataset's realism and complexity.

We utilized the M2 Apple Chip CPU on our local machine to run these experiments, which last about $30$ minutes for SVN with a full SVN-Hessian and using the anisotropic Gaussian Curvature Kernel.

\subsection{UCI Experiments}
\label{app sub sec: uci experiments}

For all the UCI experiments, we use the following hyperparameters originally based on \citet{ober2019benchmarking} for all methods: 

\begin{itemize}
    \item number of particles: $N=5$
    \item learning rate: $\text{lr} \in [1e^{-4}, 3e^{-4}, 5e^{-4}, 1e^{-3}, 3e^{-3}, 5e^{-3}, 1e^{-2}, 3e^{-2}, 5e^{-2}]$
    \item Batch size: $B \in [8, 16, 32, 64]$
    \item Hidden layer sizes: $[50,50]$
    \item Number of Epochs: 50
\end{itemize}

For all experiments, we perform $K=5$ fold cross validation and then split off an additional $20\%$ off the training set as validation data. This yields the following splits: 

\begin{table}[h!]
    \centering
    \caption{Data Split Percentages}
    \begin{tabular}{lc}
        \toprule
        Split & Percentage (\%) \\
        \midrule
        Training Set & 64 \\
        Validation Set & 16 \\
        Test Set & 20 \\
        \bottomrule
    \end{tabular}
    \label{tab:split_percentages}
\end{table}

To normalize the training and test datasets using the \texttt{StandardScaler} from \citet{scikit-learn}, we start by fitting the scaler to the training data, which calculates and stores the mean and standard deviation for each feature. After the scaler is fitted and the training data is transformed, we use the same scaler to transform the test data. The transformation of the test data uses the mean and standard deviation calculated from the training data, ensuring that both datasets are standardized based on the same criteria. This consistency is crucial for the performance of many machine learning algorithms, which require features to be on a similar scale to function properly. After this, we then split the validation data from the training data.

For the SVN algorithm, we used the full Hessian approximation, with the full SVN Hessian matrix $H^{\operatorname{SVN}}$ and the anistropic Gaussian Kernel detailed in \cref{eq: anistrpoic gaussian kernel} on all datasets except for \texttt{protein}. On the \texttt{protein} dataset, we used the KFAC Hessian approximation without the anisotropic Gaussian Kernel. 

We have included the full numerical results for the UCI regression tasks in the test NLL \cref{tab: uci regression NLL} and the test MSE \cref{tab: uci regression MSE}.  We train all the models for $50$ epochs and then load the model with the lowest validation loss for final testing. For the binary classification experiments detailed in \cref{tab: uci Classification}, we load the model from the epoch with the lowest negative log-likelihood on the validation dataset.

The \textbf{Gaussian Negative Log-Likelihood} (NLL) Loss \citep{gaussian_nll_loss} is used in regression tasks to evaluate the performance of a neural network that predicts both the mean and variance of a target variable, assuming the target follows a Gaussian distribution. The loss function is designed to handle both heteroscedastic and homoscedastic cases. For a target $y$ with predicted mean $\mu$ and predicted variance $\sigma^2$, the Gaussian NLL loss is given by:

\begin{equation}
\text{loss} = \frac{1}{2} \left( \log(\max(\sigma^2, \epsilon)) + \frac{(y - \mu)^2}{\max(\sigma^2, \epsilon)} \right) + \text{const.}
\end{equation}

Here, $\epsilon$ is a small positive value (default $10^{-6}$) used to ensure numerical stability. The constant term is omitted in our experiments. While the Gaussian NLL loss is primarily used for regression, it can also be adapted for certain classification contexts, such as probabilistic regression classification or estimating continuous class scores with uncertainty, where the network predicts mean and variance for the continuous outputs.

For the \textbf{Negative Log-Likelihood} in all our classification experiments (the UCI binary classification as well as the computer vision experiments), we follow \citet{dlr187799} and average the predicted logits over our ensemble. We then use PyTorch \citep{paszke2019pytorch} to compute the corresponding cross-entropy/NLL based on the unnormalized averaged logits and targets.

We utilized a supercomputing environment for our experiments, featuring NVIDIA A100 GPUs in its compute nodes. Depending on the dataset size, the SVN algorithm takes around $2$ hours for smaller datasets and up to $18$ hours to perform the entire cross-validation for $K=5$ splits. These times can vary between sweeps, as they depend on the learning rate and batch size.

\begin{table}[ht] 
\caption{Test negative log-likelihood on UCI regression datasets. Our proposed SVN method generally outperforms the baselines.}
\label{tab: uci regression NLL}
\centering
\begin{tabular}{@{}rrrrr@{}} \toprule
& \multicolumn{4}{c}{Test Negative Log-Likelihood $\downarrow$} \\ 
\cmidrule{2-5}
Dataset & Ensembles & WGD & SVGD & SVN \\ \midrule
autompg ($n$ = 392) & $17.53_{2.85}$ & $24.23_{2.73}$ & $14.15_{1.93}$ & $\mathbf{3.94_{0.99}}$ \\
concrete ($n$ = 1030) & $7.01_{0.82}$ & $9.13_{0.21}$ & $7.66_{1.19}$ & $\mathbf{4.77_{0.47}}$ \\
energy ($n$ = 768) & $3.44_{0.69}$ & $6.45_{0.83}$ & $2.90_{0.61}$ & $\mathbf{1.48_{0.11}}$ \\
kin8nm ($n$ = 8192) & $1.07_{0.17}$ & $1.35_{0.43}$ & $4.14_{0.68}$ & $\mathbf{-1.27_{0.16}}$ \\
naval ($n$ = 11934) & $\mathbf{-5.17_{0.09}}$ & $-4.93_{0.12}$ & $\mathbf{-5.37_{0.07}}$ & $\mathbf{-5.38_{0.20}}$ \\
power ($n$ = 9568)  & $17.42_{0.49}$ & $66.57_{13.63}$ & $14.56_{1.27}$ & $\mathbf{3.36_{0.24}}$ \\
protein ($n$ = 45730) & $18.69_{1.06}$ & $22.77_{1.28}$ & $20.83_{0.76}$ & $\mathbf{3.24_{0.21}}$ \\
wine ($n$ = 1599) & $12.24_{1.37}$ & $13.69_{2.39}$ & $14.29_{1.16}$ & $\mathbf{0.39_{0.09}}$ \\
yacht ($n$ = 308) & $1.51_{0.61}$ & $0.66_{0.18}$ & $1.10_{0.64}$ & $\mathbf{-0.36_{0.19}}$ \\ \bottomrule
\end{tabular}
\end{table}

\begin{table}[ht] 
\caption{Test MSE on UCI Regression Tasks}
\label{tab: uci regression MSE}
\centering
\begin{tabular}{@{}rrrrr@{}}\toprule
& \multicolumn{4}{c}{Test MSE $\downarrow$} \\ 
\cmidrule{2-5}
Dataset & Ensembles & WGD & SVGD & SVN \\ \midrule
autompg ($n$ = 392) & $2.53_{0.04}$ & $\mathbf{0.77_{0.28}}$ & $3.22_{0.15}$ & $1.61_{0.28}$ \\
concrete ($n$ = 1030) & $\mathbf{2.68_{0.66}}$ & $\mathbf{2.61_{0.31}}$ & $\mathbf{2.42_{0.81}}$ & $\mathbf{2.45_{0.72}}$ \\
energy ($n$ = 768) & $0.19_{0.04}$ & $0.19_{0.05}$ & $0.14_{0.02}$ & $\mathbf{0.07_{0.01}}$ \\
kin8nm ($n$ = 8192) & $\mathbf{7.4e^{-6} \pm 2.36 e^{-6}}$ & $4.5 e^{-5} \pm 1.3e^{-5}$ & $1.9e^{-5} \pm 0.3 e^{-5}$ & $7.06 e^{-5} \pm 1.42 e^{-5}$ \\
naval ($n$ = 11934) & $\mathbf{6.2e^{-8} \pm 1.4e^{-8}}$ & $5.4e^{-7} \pm 1.2e^{-7}$ & $2.8e^{-7} \pm 0.7e^{-7}$ & $1.2e^{-7} \pm 0.3e^{-7}$ \\
power ($n$ = 9568)  & $0.16_{0.02}$ & $0.49_{0.09}$ & $0.77_{0.06}$ & $\mathbf{0.08_{0.02}}$ \\
protein ($n$ = 45730) & $\mathbf{0.11_{0.11}}$ & $\mathbf{0.06_{0.01}}$ & $\mathbf{0.11_{0.11}}$ & $0.36_{0.10}$ \\
wine ($n$ = 1599) & $0.05_{0.01}$ & $\mathbf{0.02_{0.01}}$ & $\mathbf{0.04_{0.01}}$ & $\mathbf{0.02_{0.01}}$ \\
yacht ($n$ = 308) & $\mathbf{0.51_{0.20}}$ & $\mathbf{0.26_{0.08}}$ & $0.63_{0.10}$ & $\mathbf{0.61_{0.47}}$ \\ \bottomrule
\end{tabular}
\end{table}

\subsection{Computer Vision Experiments}
\label{app sub sec: cv experiments}

For our computer vision experiments on the MNIST \citep{lecun1998mnist} and FashionMNIST datasets \citep{xiao2017fashion}, we performed cross-validation with $K=5$. We then split the validation data as described in \cref{app sub sec: uci experiments} to obtain the final splits, as detailed in \cref{tab:split_percentages}.

\begin{itemize}
    \item number of particles: $N=5$
    \item learning rate: $\text{lr} \in [1e^{-3}, 3e^{-3}, 5e^{-3}, 1e^{-2}, 3e^{-2}]$
    \item learning rate schedule: constant
    \item Batch size: $B \in [64, 128, 256]$
    \item Optimizer: Adam 
    \item Number of Epochs SVN: 20
    \item Number of Epochs LL-SVN: 100
    \item Number of Epochs SVGD: 100
    \item Number of Epochs Ensemble: 100
\end{itemize}

For all our experiments, we used the LenNet architecture \citep{lenet}. We also experimented with weight decay and gradient clipping but found no improvement. For our SVN algorithm, we used the Diagonal Hessian Approximation as well as the Block Diagonal Approximation in conjunction with the anisotropic curvature kernel to ensure the method scales well to larger models. We limited the number of epochs to evaluate its ability to converge faster than the other methods and to decrease the computational requirements of the experiments. For LL-SVN, we used the KFAC Hessian approximation for the parameters of the last layer.

We utilized a supercomputing environment for our experiments, with NVIDIA A100 GPUs in its compute nodes. The LL-SVN algorithm takes about 2 -- 3 hours to finish training. SVN requires around $16$ hours for the MNIST dataset and around $25$ hours for running 20 epochs on the MNIST and FashionMNIST datasets, respectively. These numbers can vary depending on the batch size used and the learning rate.

\section{Further Experiments}

\subsection{SVN in low-data regime}

\begin{figure}
    \centering    
    \includegraphics[width=0.7\textwidth]{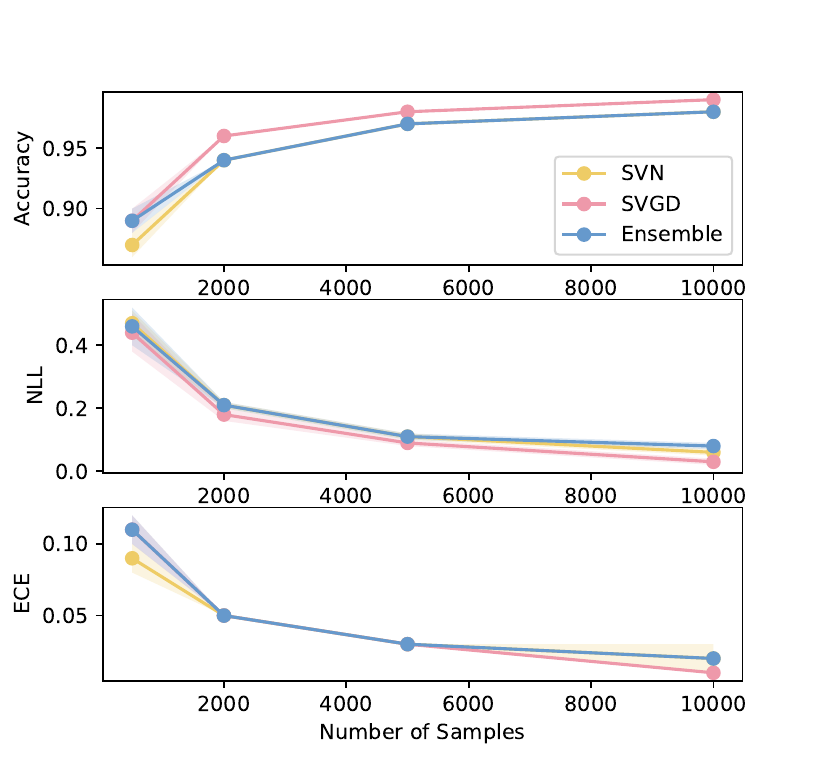}
    \caption{Performance metrics for Ensemble, SVGD, and SVN on subsampled MNIST data with varying sizes ($n \in \{500, 2000, 5000, 10000\}$). The plots display accuracy, negative log-likelihood, and expected calibration error.}
    \label{fig: subsample}
\end{figure}

To further investigate the performance of SVN ensembles in low-data scenarios, we conducted an additional evaluation using subsamples of the MNIST dataset \citep{lecun1998mnist}. By employing a random seed, we created subsampled datasets with sizes $n \in \{500, 2000, 5000, 10000\}$. The results of this evaluation are depicted in \cref{fig: subsample}, where we present the accuracy, negative log-likelihood, and expected calibration error for three methods: Ensemble, Stein Variational Gradient Descent (SVGD), and SVN.

Our analysis revealed that while SVGD consistently achieved higher accuracy across all subsample sizes, SVN demonstrated superior calibration performance, particularly in the low-data regime. Despite these differences, the overall performance of the three methods remained comparable under the conditions tested. This suggests that both SVN and SVGD offer a competitive alternative to the Ensemble in scenarios with limited data availability.

\subsection{Ablation Studies: }
\label{app sub sec: ablation}

An overview of SVN's main hyperparameters are: 

\begin{itemize}
    \item Choice of Hessian Approximation
    \item Anistropic Gaussian Curvature Kernel
    \item Use of the Block Diagonal Approximation
    \item Number of particles $N$
\end{itemize}

\paragraph{Block Diagonal Approximation and Curvature Kernel}

\begin{figure}
    \centering
    \includegraphics[width=0.6\textwidth]{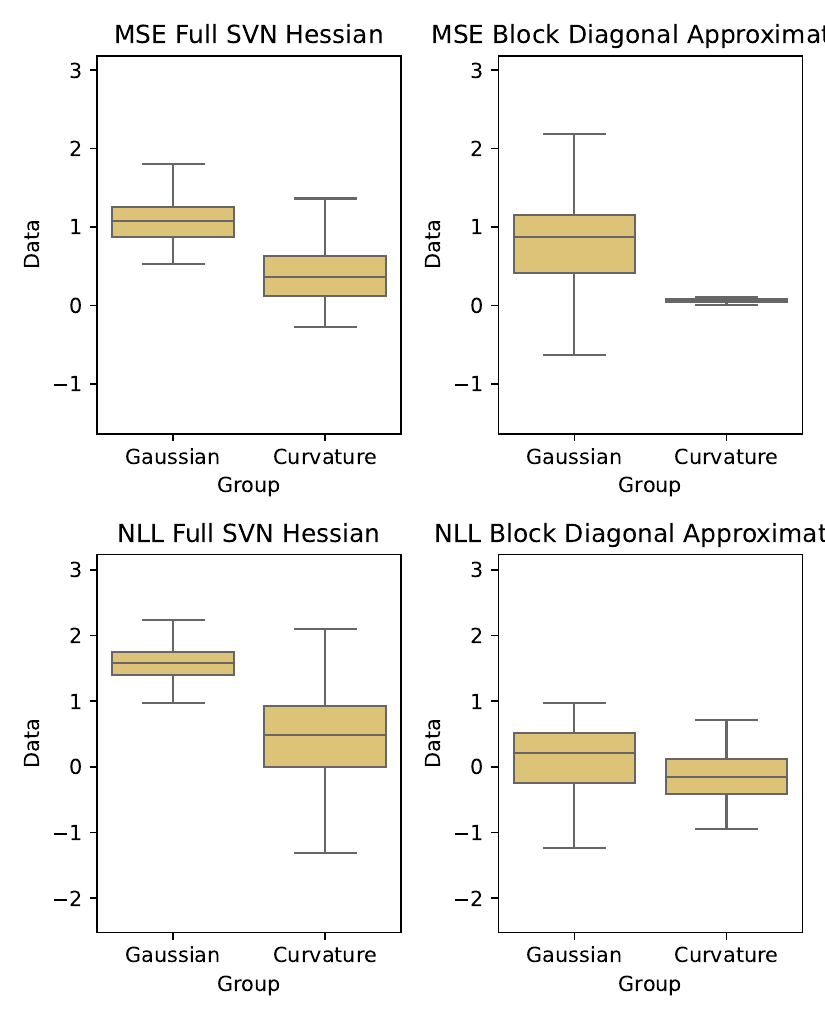}
    \caption{Ablation study on the use of the Block Diagonal Approximation and Anistropic Curvature Kernel. The first and third plot correpsond to the use of the Full SVN in computing the SVN update, while the second and fourth plot show the Block Diagonal approximation.}
    \label{fig:ablation_curvature_and_block_diagonal_distribution}
\end{figure}

\Cref{fig:ablation_curvature_and_block_diagonal_distribution} demonstrates several key findings regarding the performance and stability of different kernel choices in the SVN algorithm for the Yacht dataset. Firstly, using the anisotropic Gaussian kernel or the curvature kernel appears to improve performance in terms of mean squared error (MSE) and stabilize training, as indicated by smaller standard errors. Secondly, with regard to the negative log-likelihood (NLL) metric, the curvature kernel interestingly makes a more significant impact on improving NLL when using the full SVN Hessian. In contrast, when the block-diagonal approximation is employed, the anisotropic Gaussian kernel slightly improves performance.

\paragraph{Number of Particles}

\begin{figure}[h]
    \centering
    \includegraphics[width=0.7\linewidth]{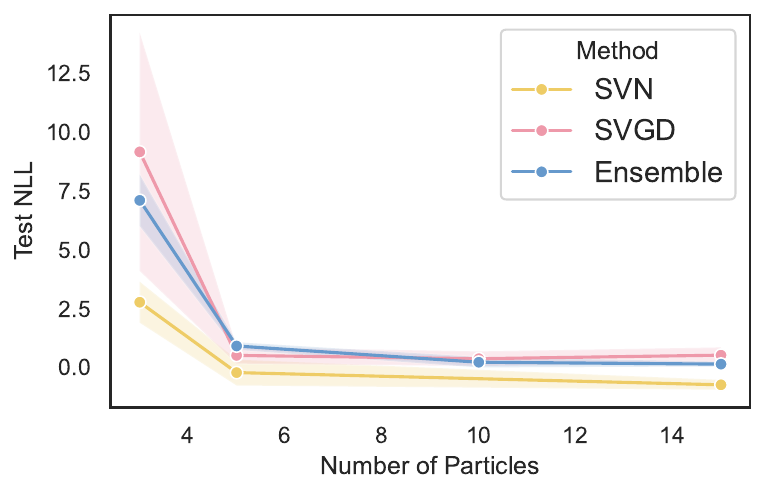}
    \caption{Test NLL for methods with varying numbers of particles for the Yacht dataset. While SVN exhibits superior performance for all considered numbers of particles, it is the only method with a consistent downwards trend, in contrast to SVGD and Ensembles, which appear to flatten for larger numbers of particles.}
    \label{fig:ablation_number_of_particles}
\end{figure}

Since an important hyperparameter in particle-based variational inference (VI) is the number of particles $N$, we investigated the behavior of the methods with different numbers of particles. As depicted in \cref{fig:ablation_number_of_particles}, the results for the Yacht dataset provide significant insights into the impact of particle quantity on performance.

It can be observed that all methods generally improve with an increasing number of particles, as one might expect. However, while the test NLL of Ensembles and SVGD appears to plateau with larger numbers of particles, SVN demonstrates a more consistent improvement trend. This suggests that SVN's performance continues to benefit from additional particles without encountering the diminishing returns seen in the other methods.

We hypothesize that the curvature information incorporated in SVN updates allows it to more effectively allocate particles within the posterior landscape. This capability likely enables SVN to scale better with a larger number of particles, maintaining its performance improvement. In contrast, Ensembles may suffer from mode collapse, where particles fail to adequately explore the posterior distribution beyond a certain point. SVGD, while more robust than Ensembles, also appears to reach a saturation point where additional particles do not translate into significant performance gains.

These findings underscore the potential advantages of SVN in scenarios requiring the deployment of numerous particles, highlighting its robustness and scalability. Further research could investigate the underlying mechanisms in more detail and explore ways to mitigate the mode collapse observed in Ensemble methods \citep{pop2018deep}.

\subsection{Hessian Evolution during SVN Training}
\label{app sub sec: hessian plots}

We aim to gain deeper insights into the behavior of our method, particularly in light of the observation in \cref{sec: Experiments} that SVN exhibits considerably faster convergence of the validation NLL scores during training compared to SVGD or Ensembles. However, during the first one to two epochs its performance is considerably worse. Thus, we conducted additional experiments on a scaled-down architecture. The aim is to visualize the Hessian matrix throughout the training process. For this purpose, we trained the model on the Yacht dataset from the UCI regression benchmark datasets. The following hyperparameters are used for the experiment: number of particles: $N=5$, learning rate: $\text{lr} = 1e^{-2}$, batch size: $B = 16$, single hidden fully connected layer of size: $10$. This results in a total of 81 trainable parameters of the model. 

We utilized full Hessian approximations for the gradient update using the Laplace library \citep{daxberger2022laplace}. The results were evaluated on a 20\% holdout of unseen data. The training was conducted over 100 epochs, and snapshots of the Hessian were taken at six different points during the training process. The experiment converged to a test MSE score of $0.164$ and a test NLL score of $0.382$.

\begin{figure}[H]
\centering
    \begin{subfigure}[h]{0.3\textwidth}
         \centering
         \includegraphics[width=\textwidth]{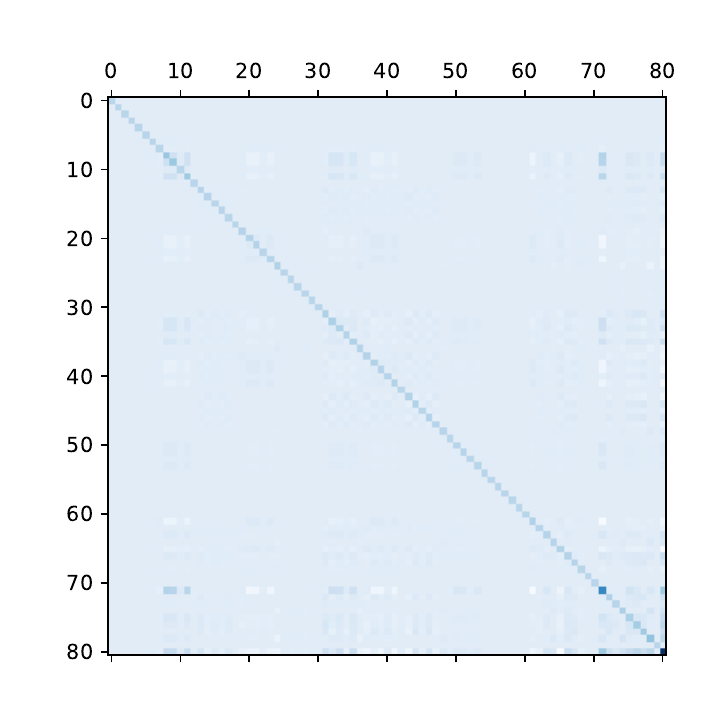}
         \caption{end of epoch 1}
    \end{subfigure}
    \begin{subfigure}[h]{0.3\textwidth}
         \centering
         \includegraphics[width=\textwidth]{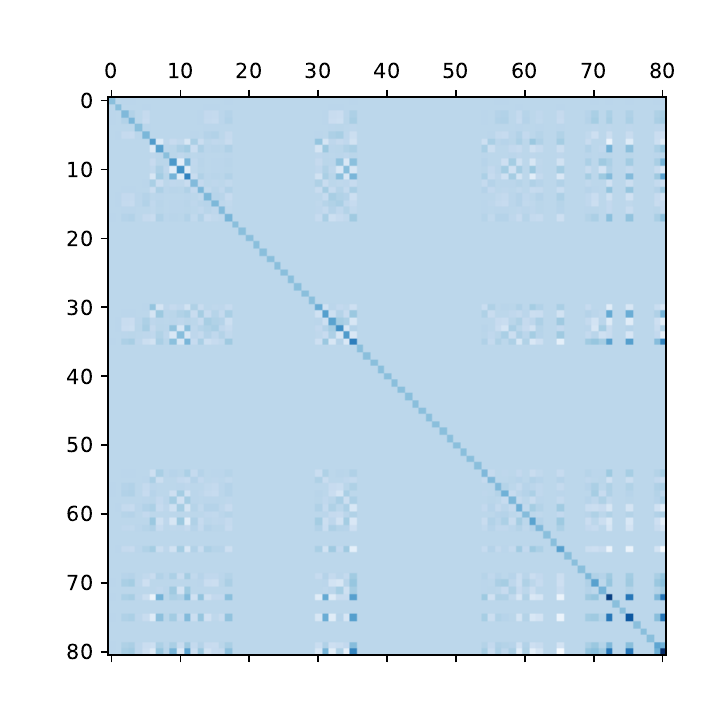}
         \caption{end of epoch 20}
    \end{subfigure} 
    \begin{subfigure}[h]{0.3\textwidth}
         \centering
         \includegraphics[width=\textwidth]{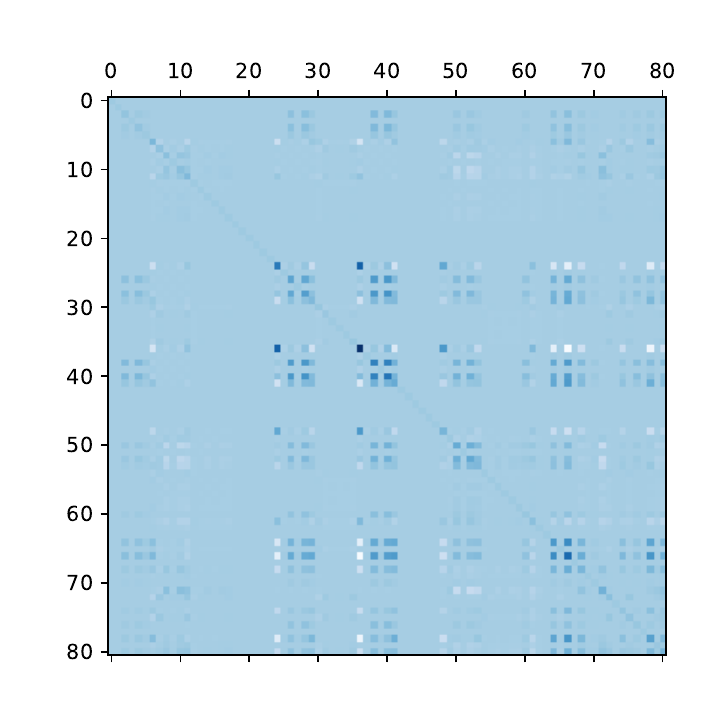}
         \caption{end of epoch 40}
    \end{subfigure} 
    \begin{subfigure}[h]{0.3\textwidth}
         \centering
         \includegraphics[width=\textwidth]{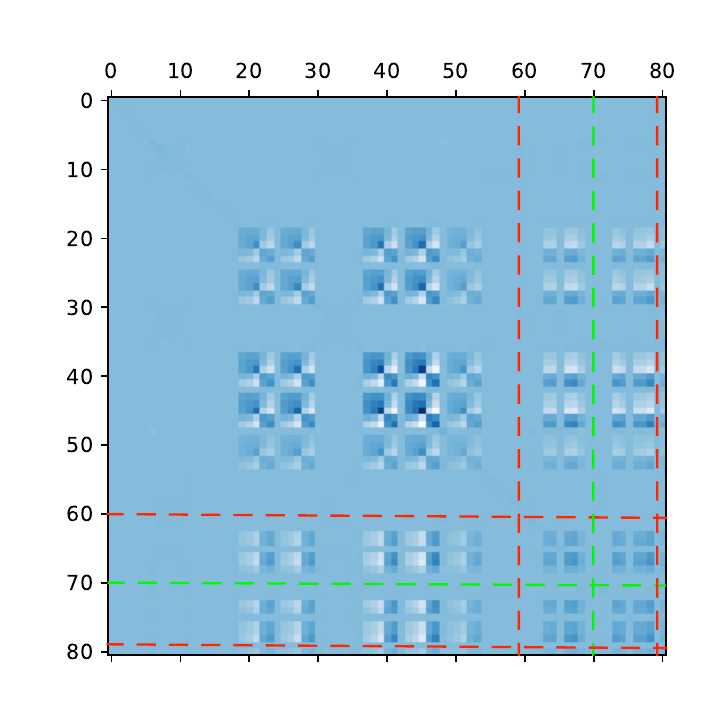}
         \caption{end of epoch 60}
    \end{subfigure} 
    \begin{subfigure}[h]{0.3\textwidth}
         \centering
         \includegraphics[width=\textwidth]{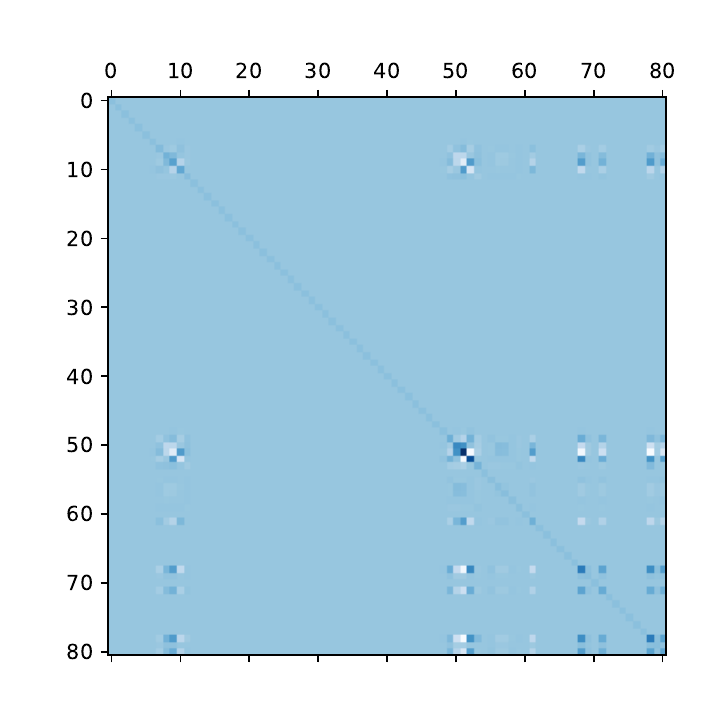}
         \caption{end of epoch 80}
    \end{subfigure} 
    \begin{subfigure}[h]{0.3\textwidth}
         \centering
         \includegraphics[width=\textwidth]{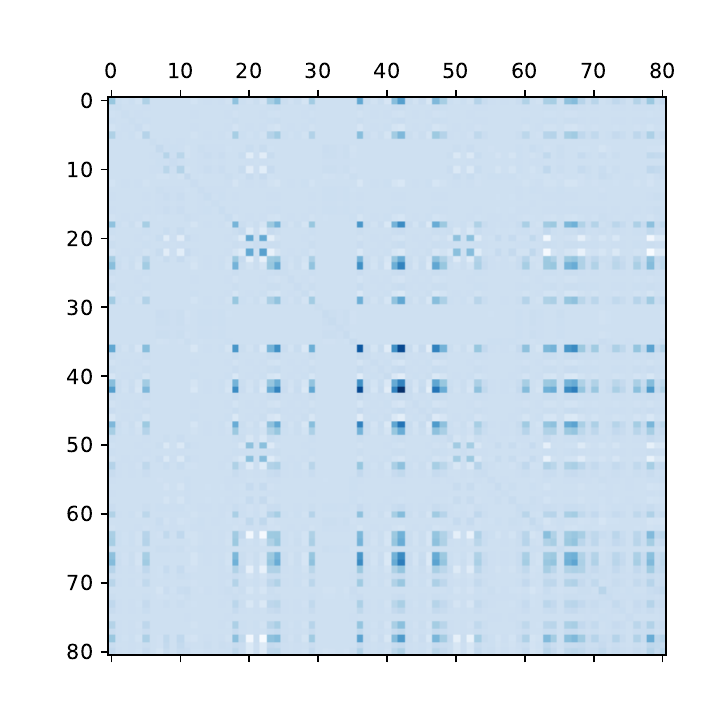}
         \caption{end of epoch 100}
    \end{subfigure} 
\centering
\caption{Visualization of snapshots of Hessian approximations for \textbf{Particle 1} at different times during the training process. Furthermore, the drawn lines in (d) illustrate the separation of weights and biases of different layers. The top left area for axes [0,59] represent the weight connections of input layer with the hidden layer, the next range [60,69] represent bias terms of the first layer, the next range [70,79] represent weight connections from hidden layer to output layer and the axes at 80 represent the bias term for output neuron.}
\label{fig:hess1}
\end{figure}

\begin{figure}[H]
\centering
    \begin{subfigure}[h]{0.3\textwidth}
         \centering
         \includegraphics[width=\textwidth]{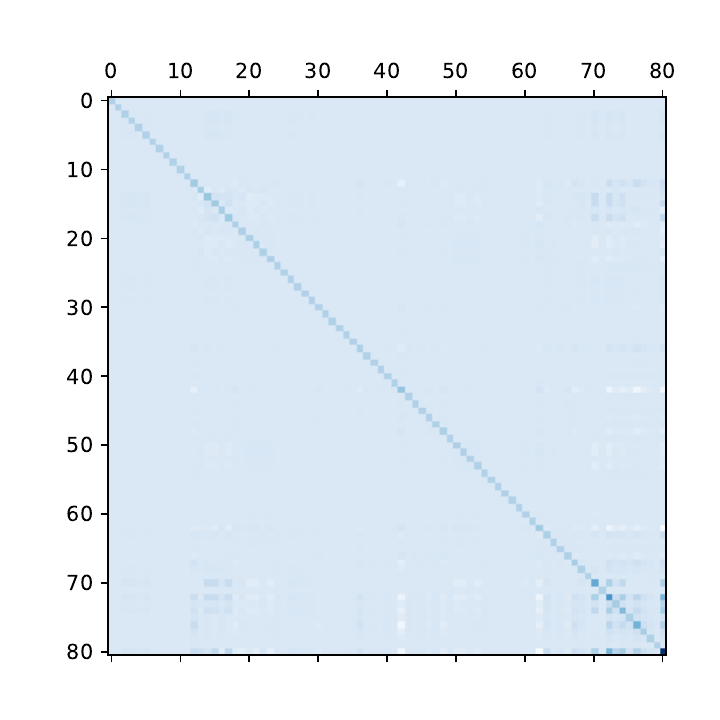}
         \caption{end of epoch 1}
    \end{subfigure} 
    \begin{subfigure}[h]{0.3\textwidth}
         \centering
         \includegraphics[width=\textwidth]{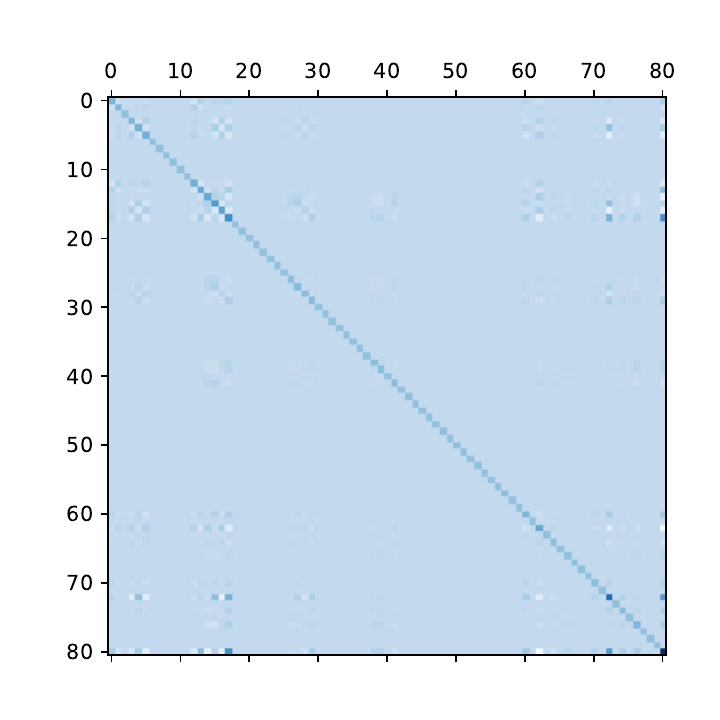}
         \caption{end of epoch 20}
    \end{subfigure} 
    \begin{subfigure}[h]{0.3\textwidth}
         \centering
         \includegraphics[width=\textwidth]{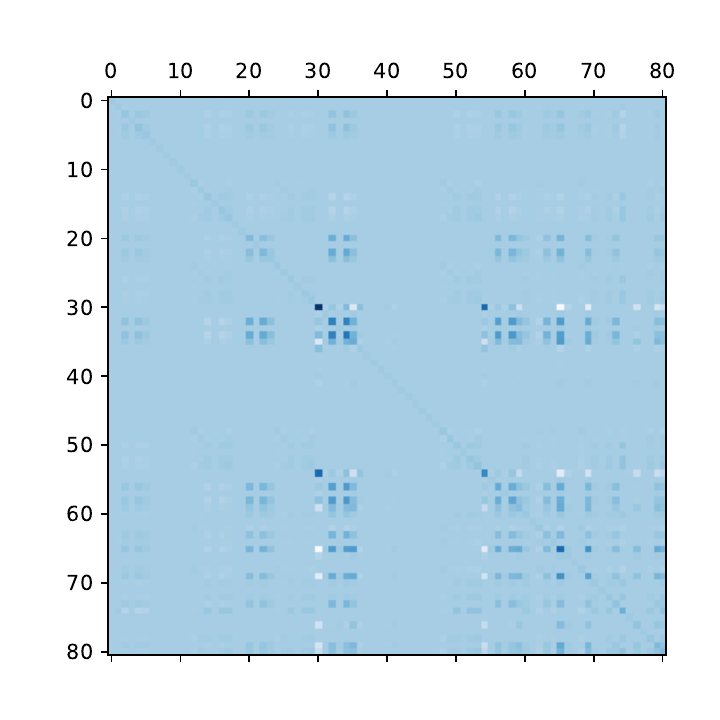}
         \caption{end of epoch 40}
    \end{subfigure} 
    
    \begin{subfigure}[h]{0.3\textwidth}
         \centering
         \includegraphics[width=\textwidth]{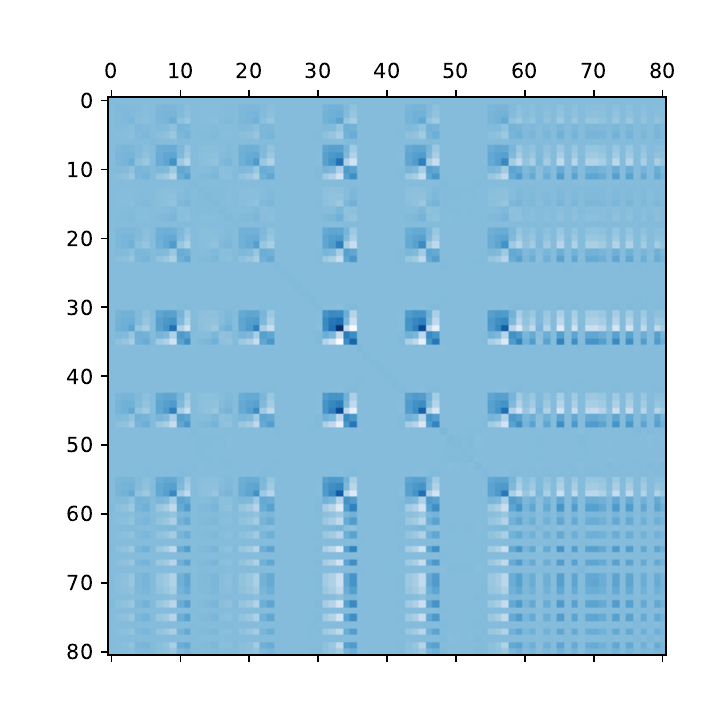}
         \caption{end of epoch 60}
    \end{subfigure} 
    \begin{subfigure}[h]{0.3\textwidth}
         \centering
         \includegraphics[width=\textwidth]{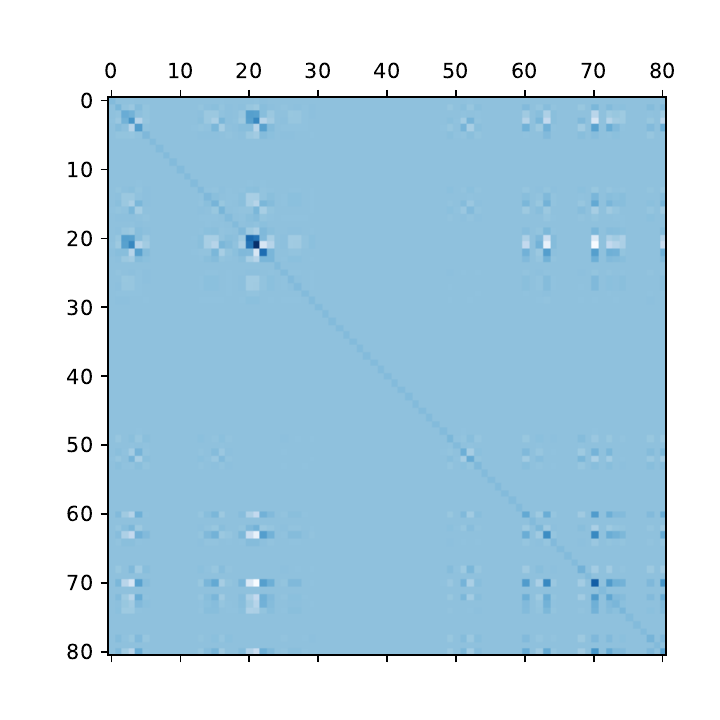}
         \caption{end of epoch 80}
    \end{subfigure} 
    \begin{subfigure}[h]{0.3\textwidth}
         \centering
         \includegraphics[width=\textwidth]{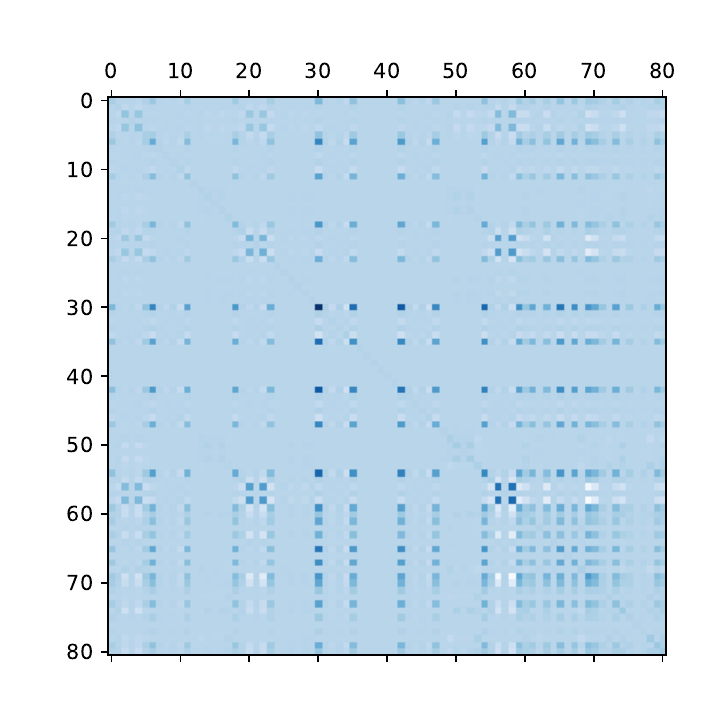}
         \caption{end of epoch 100}
    \end{subfigure} 
\centering
\caption{Visualization of snapshots of Hessian approximations for \textbf{Particle 2} at different times during the training process.}
\label{fig:hess2}
\end{figure}

\begin{figure}[H]
\centering
    \begin{subfigure}[h]{0.3\textwidth}
         \centering
         \includegraphics[width=\textwidth]{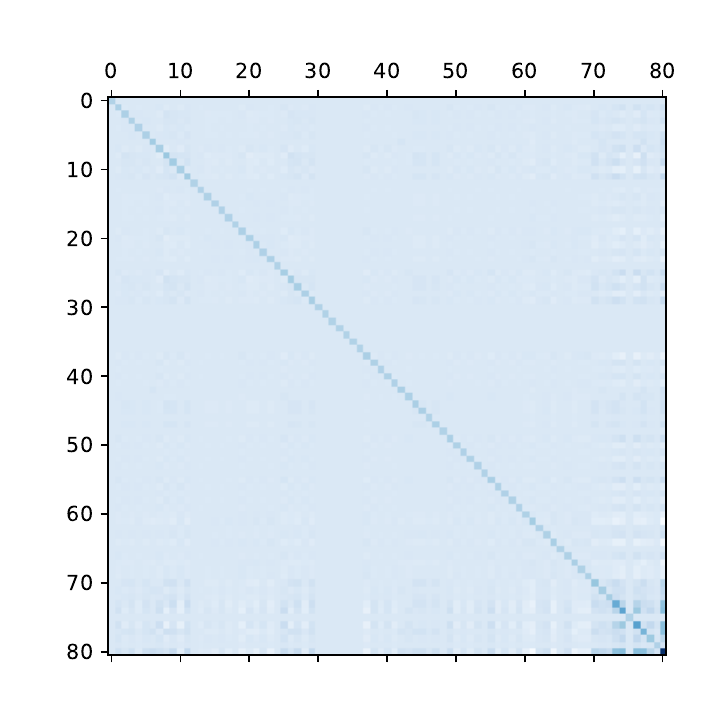}
         \caption{end of epoch 1}
    \end{subfigure} 
    \begin{subfigure}[h]{0.3\textwidth}
         \centering
         \includegraphics[width=\textwidth]{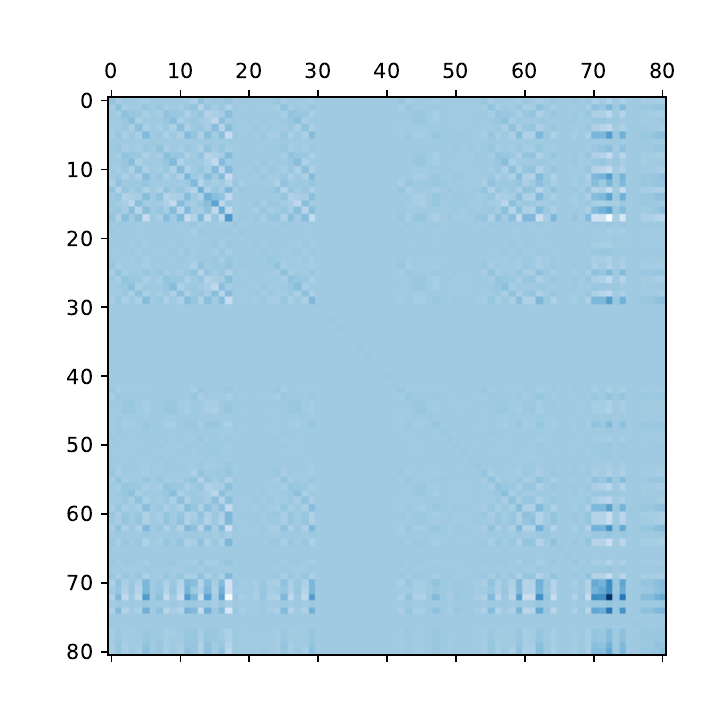}
         \caption{end of epoch 20}
    \end{subfigure} 
    \begin{subfigure}[h]{0.3\textwidth}
         \centering
         \includegraphics[width=\textwidth]{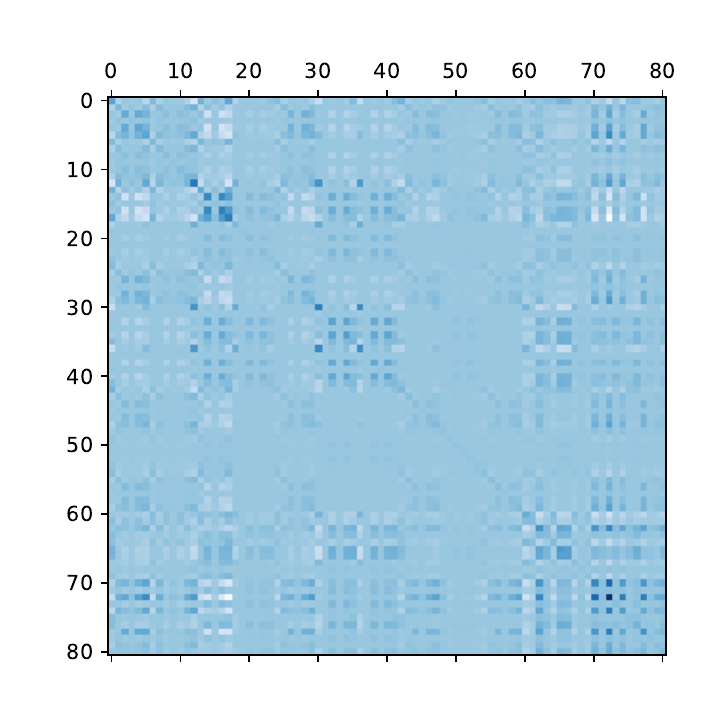}
         \caption{end of epoch 40}
    \end{subfigure} 
    
    \begin{subfigure}[h]{0.3\textwidth}
         \centering
         \includegraphics[width=\textwidth]{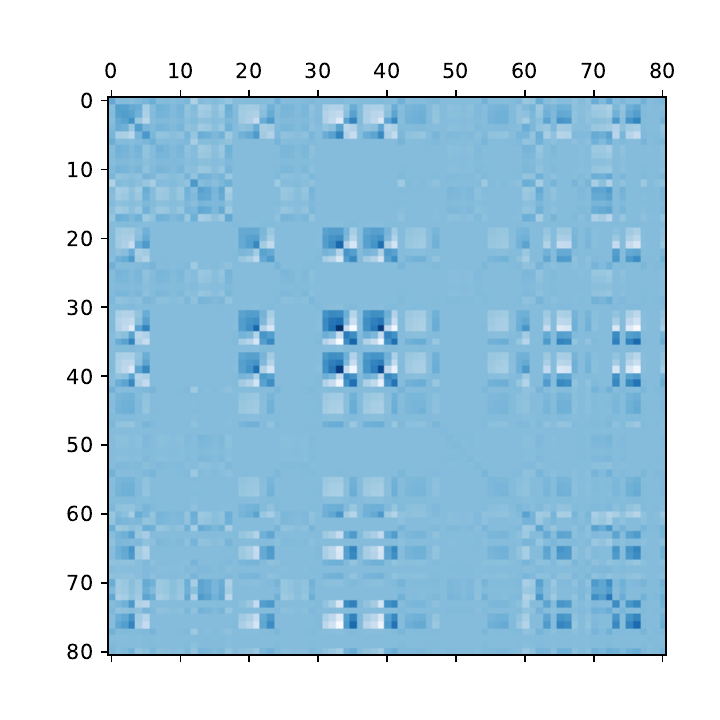}
         \caption{end of epoch 60}
    \end{subfigure} 
    \begin{subfigure}[h]{0.3\textwidth}
         \centering
         \includegraphics[width=\textwidth]{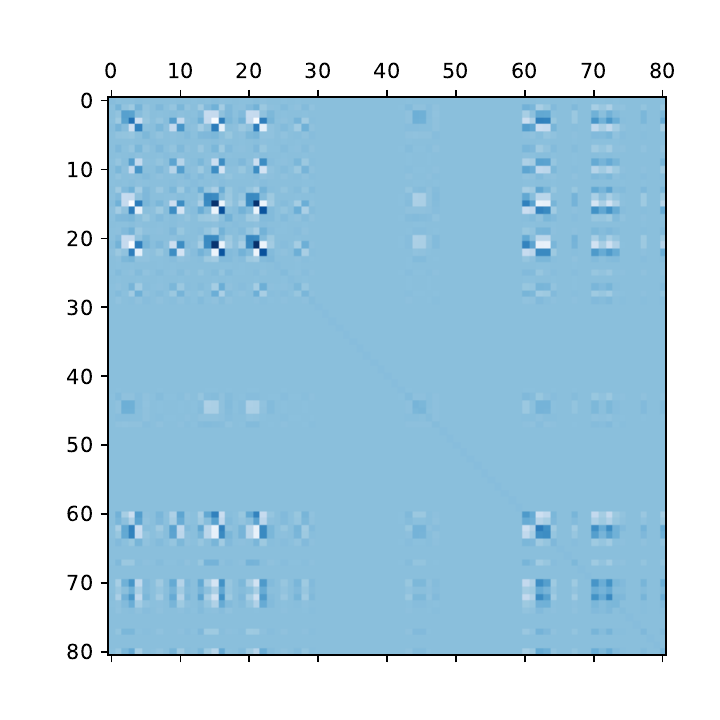}
         \caption{end of epoch 80}
    \end{subfigure} 
    \begin{subfigure}[h]{0.3\textwidth}
         \centering
         \includegraphics[width=\textwidth]{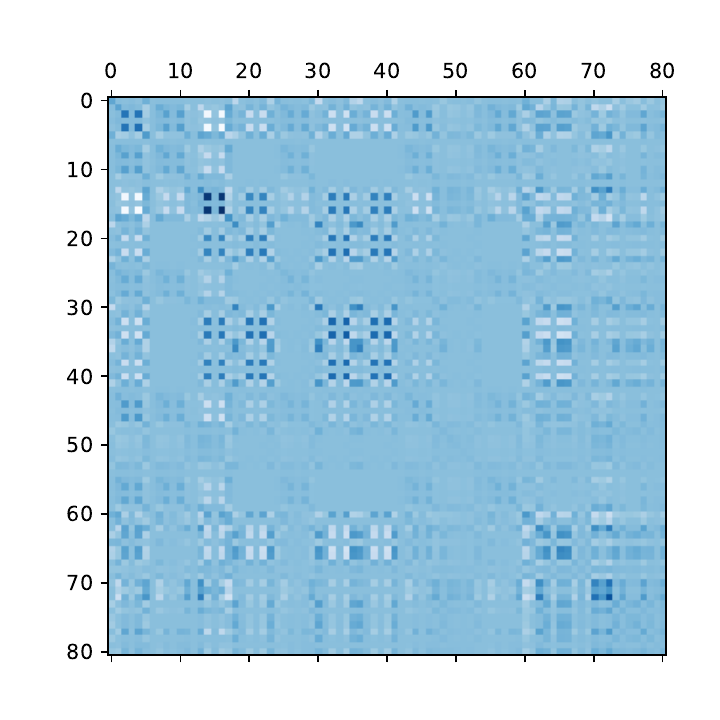}
         \caption{end of epoch 100}
    \end{subfigure}

\centering
\caption{Visualization of snapshots of Hessian approximations for \textbf{Particle 3} at different times during the training process.}
\label{fig:hess3}
\end{figure}

\begin{figure}[H]
\centering
    \begin{subfigure}[h]{0.3\textwidth}
         \centering
         \includegraphics[width=\textwidth]{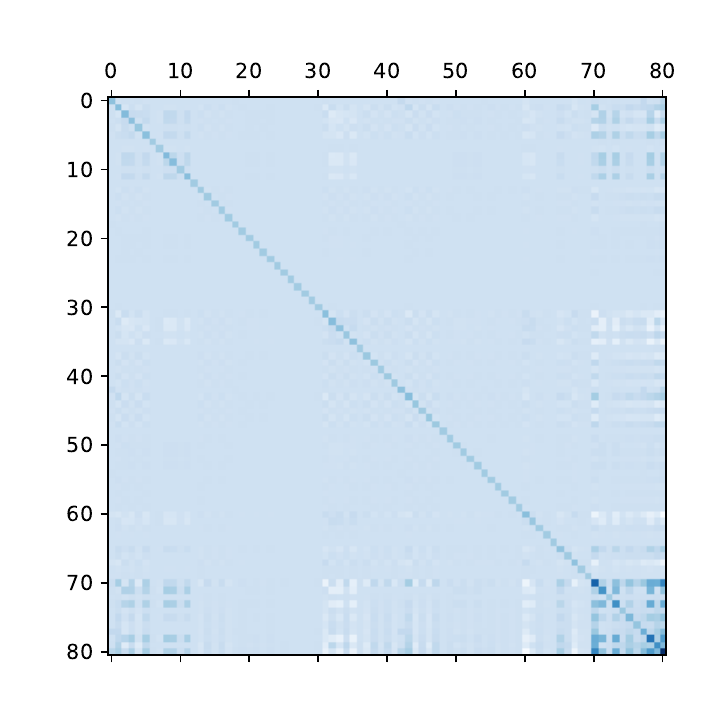}
         \caption{end of epoch 1}
    \end{subfigure} 
    \begin{subfigure}[h]{0.3\textwidth}
         \centering
         \includegraphics[width=\textwidth]{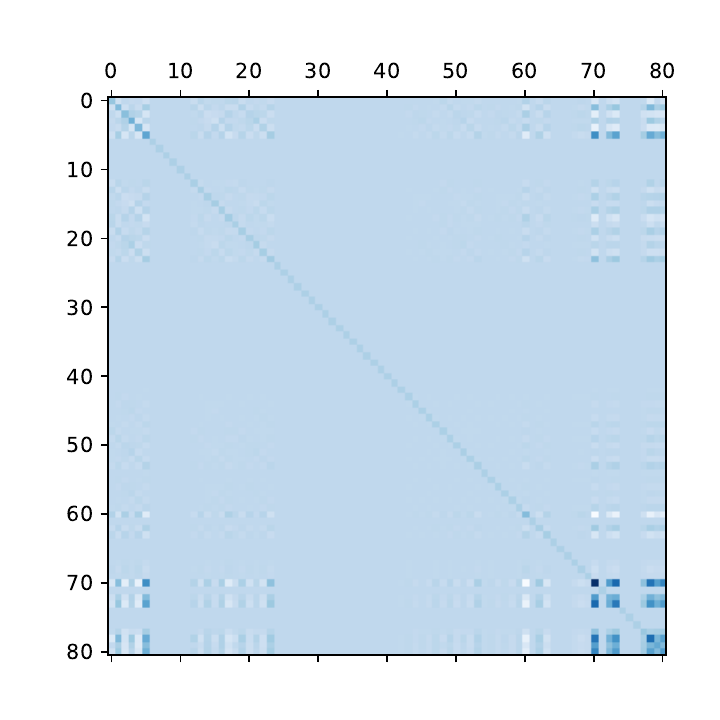}
         \caption{end of epoch 20}
    \end{subfigure} 
    \begin{subfigure}[h]{0.3\textwidth}
         \centering
         \includegraphics[width=\textwidth]{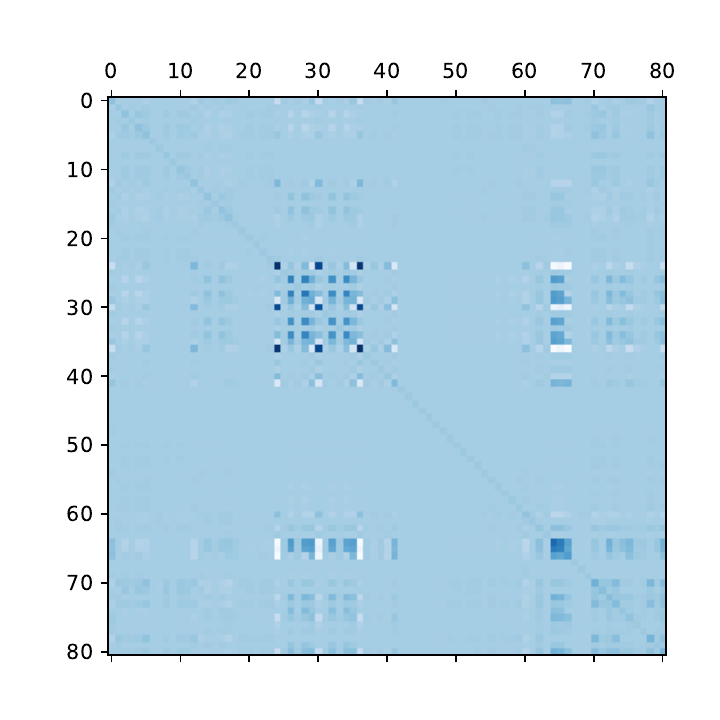}
         \caption{end of epoch 40}
    \end{subfigure} 
    
    \begin{subfigure}[h]{0.3\textwidth}
         \centering
         \includegraphics[width=\textwidth]{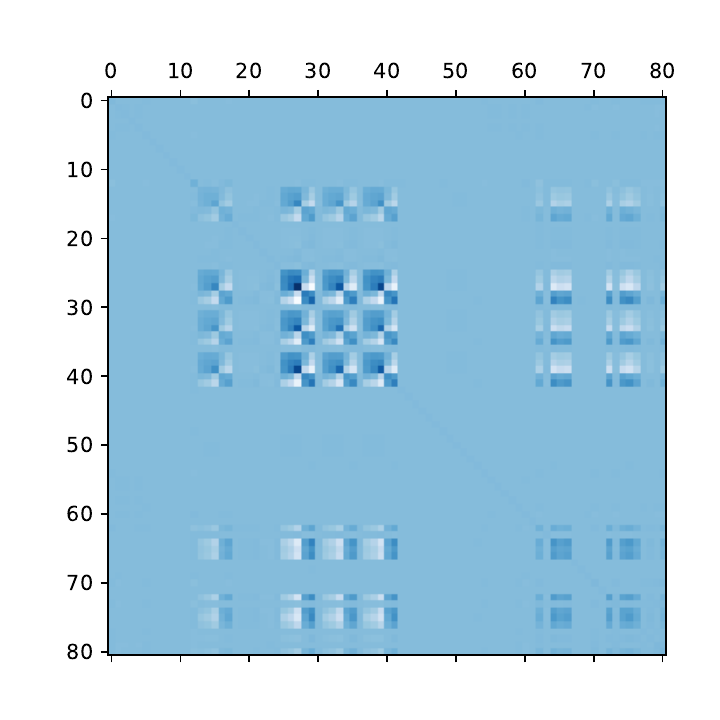}
         \caption{end of epoch 60}
    \end{subfigure} 
    \begin{subfigure}[h]{0.3\textwidth}
         \centering
         \includegraphics[width=\textwidth]{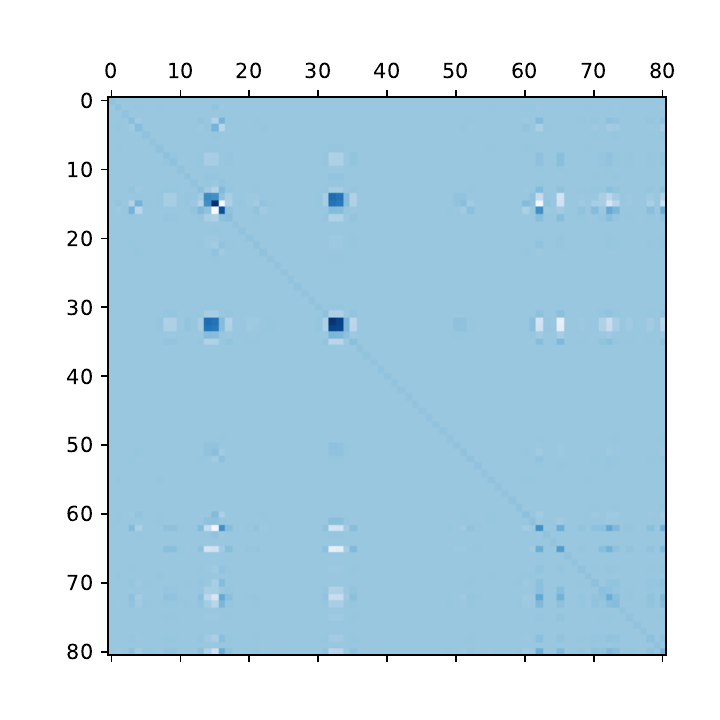}
         \caption{end of epoch 80}
    \end{subfigure} 
    \begin{subfigure}[h]{0.3\textwidth}
         \centering
         \includegraphics[width=\textwidth]{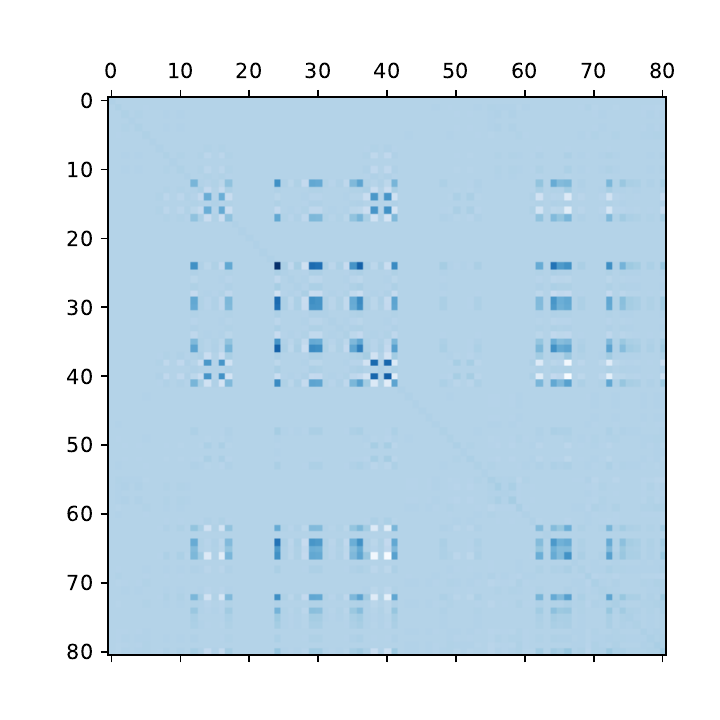}
         \caption{end of epoch 100}
    \end{subfigure}

\centering
\caption{Visualization of snapshots of Hessian approximations for \textbf{Particle 4} at different times during the training process.}
\label{fig:hess4}
\end{figure}

\begin{figure}[H]
\centering
    \begin{subfigure}[h]{0.3\textwidth}
         \centering
         \includegraphics[width=\textwidth]{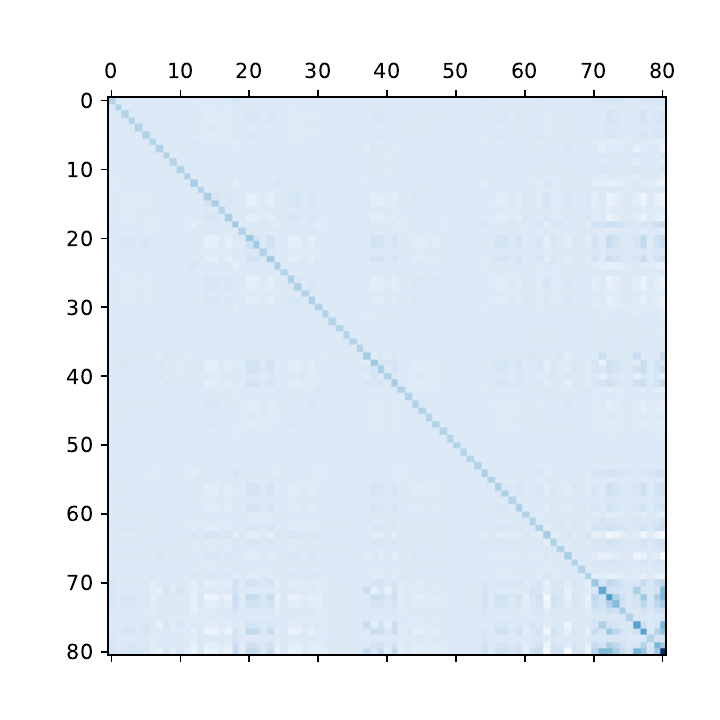}
         \caption{end of epoch 1}
    \end{subfigure} 
    \begin{subfigure}[h]{0.3\textwidth}
         \centering
         \includegraphics[width=\textwidth]{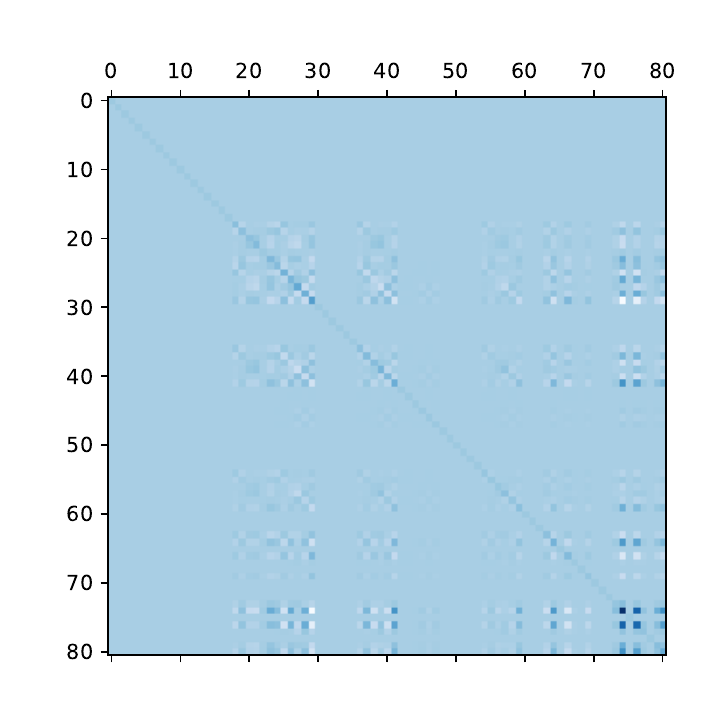}
         \caption{end of epoch 20}
    \end{subfigure} 
    \begin{subfigure}[h]{0.3\textwidth}
         \centering
         \includegraphics[width=\textwidth]{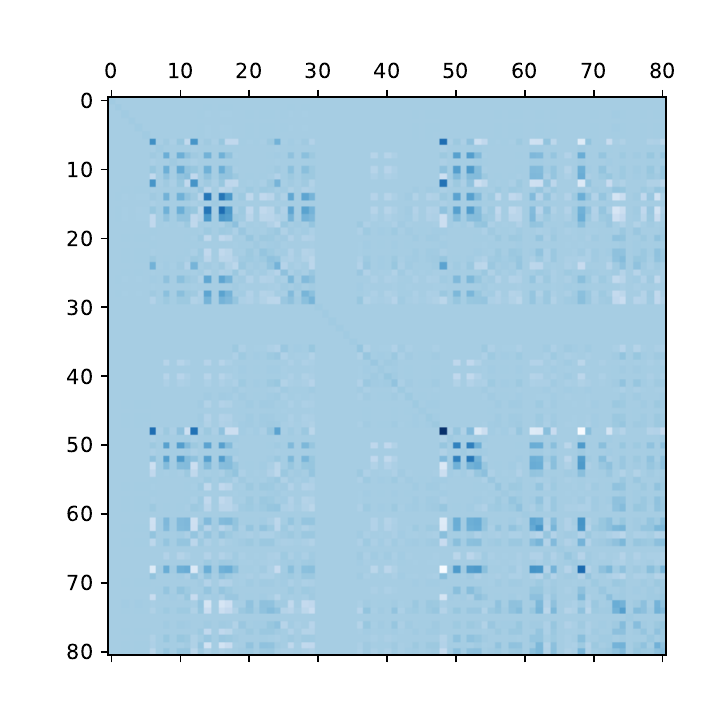}
         \caption{end of epoch 40}
    \end{subfigure} 
    
    \begin{subfigure}[h]{0.3\textwidth}
         \centering
         \includegraphics[width=\textwidth]{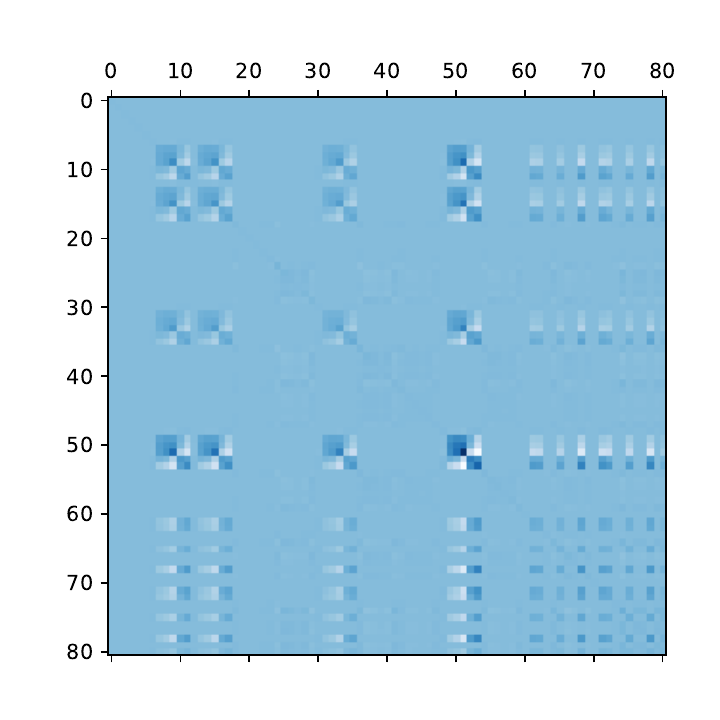}
         \caption{end of epoch 60}
    \end{subfigure} 
    \begin{subfigure}[h]{0.3\textwidth}
         \centering
         \includegraphics[width=\textwidth]{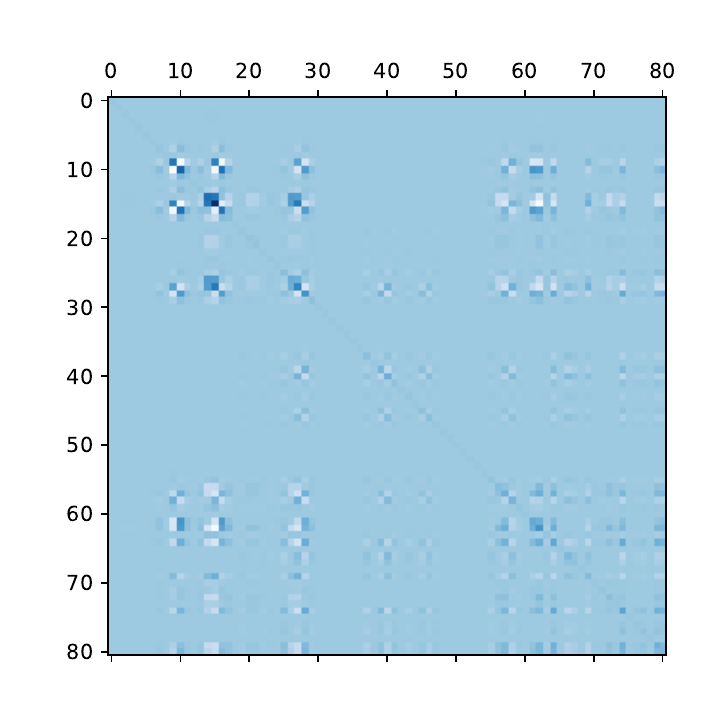}
         \caption{end of epoch 80}
    \end{subfigure} 
    \begin{subfigure}[h]{0.3\textwidth}
         \centering
         \includegraphics[width=\textwidth]{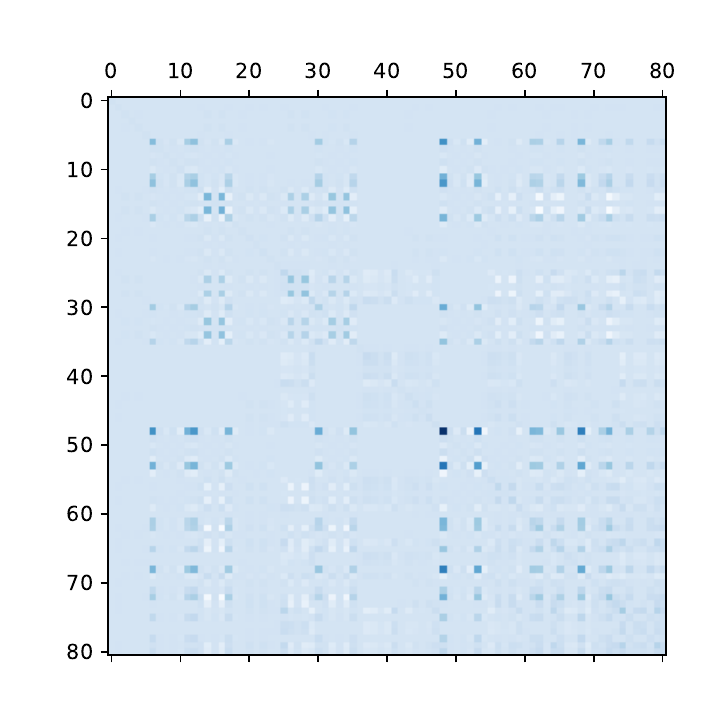}
         \caption{end of epoch 100}
    \end{subfigure} 
    
\centering
\caption{Visualization of snapshots of Hessian approximations for \textbf{Particle 5} at different times during the training process.}
\label{fig:hess5}
\end{figure}

\begin{figure}[H]
\centering
    \begin{subfigure}[h]{0.3\textwidth}
         \centering
         \includegraphics[width=\textwidth]{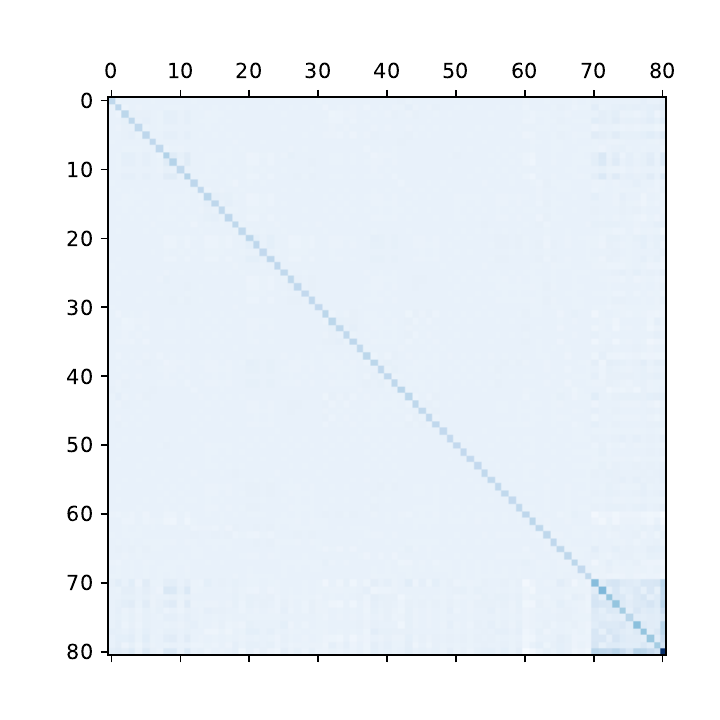}
         \caption{end of epoch 1}
    \end{subfigure} 
    \begin{subfigure}[h]{0.3\textwidth}
         \centering
         \includegraphics[width=\textwidth]{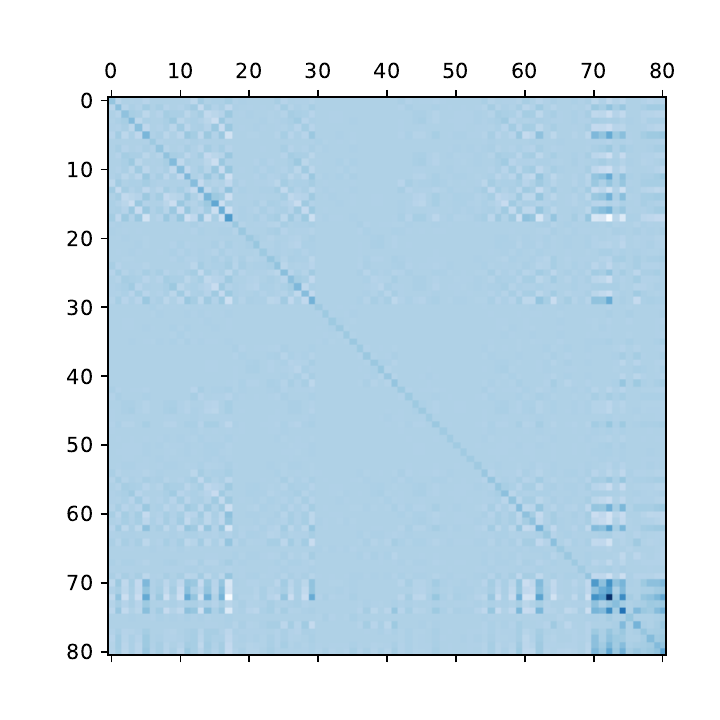}
         \caption{end of epoch 20}
    \end{subfigure} 
    \begin{subfigure}[h]{0.3\textwidth}
         \centering
         \includegraphics[width=\textwidth]{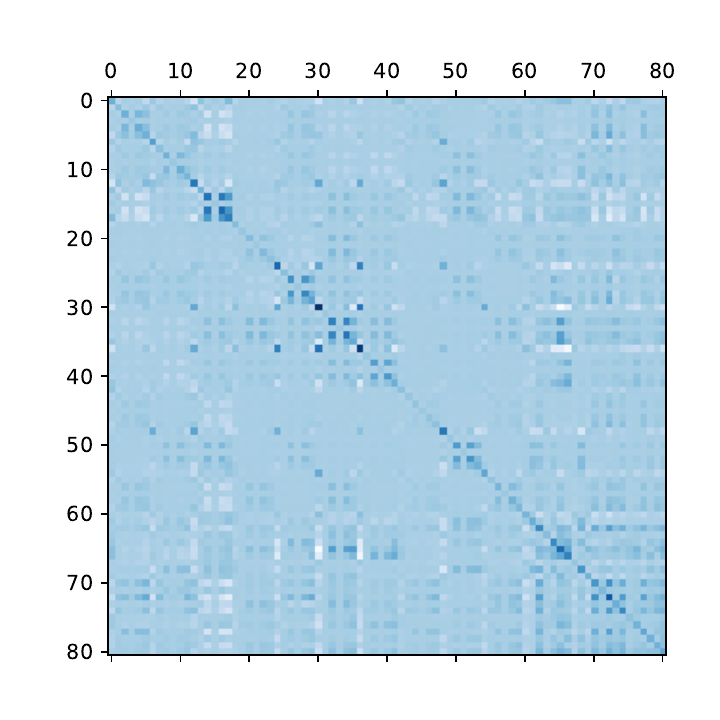}
         \caption{end of epoch 40}
    \end{subfigure} 
    
    \begin{subfigure}[h]{0.3\textwidth}
         \centering
         \includegraphics[width=\textwidth]{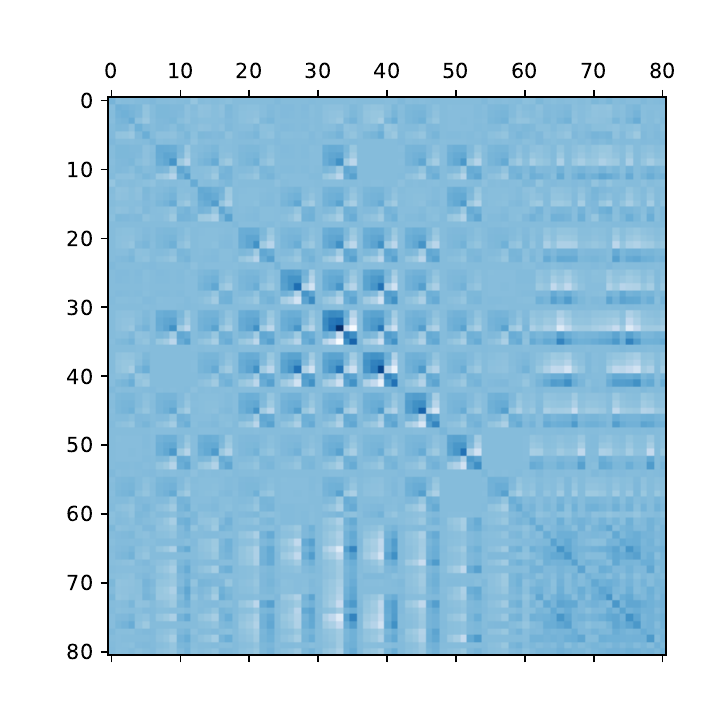}
         \caption{end of epoch 60}
    \end{subfigure} 
    \begin{subfigure}[h]{0.3\textwidth}
         \centering
         \includegraphics[width=\textwidth]{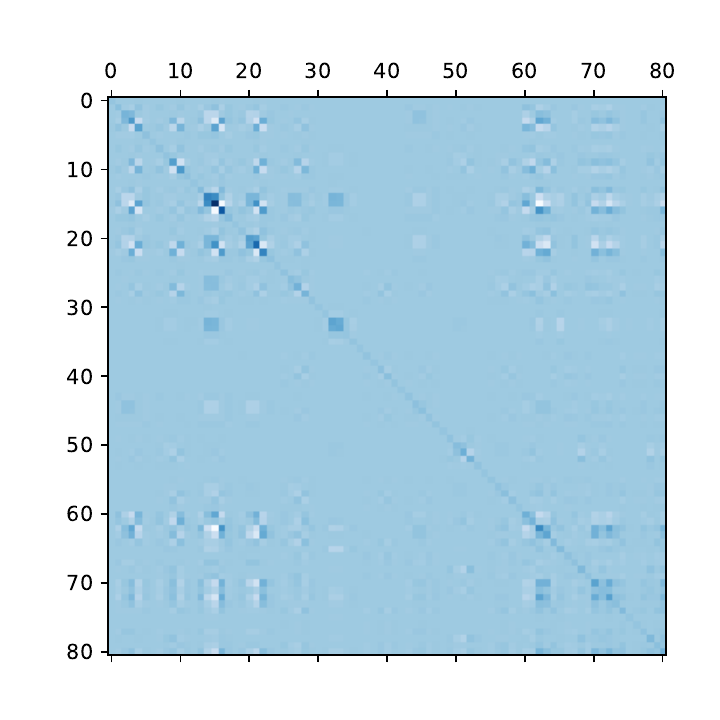}
         \caption{end of epoch 80}
    \end{subfigure} 
    \begin{subfigure}[h]{0.3\textwidth}
         \centering
         \includegraphics[width=\textwidth]{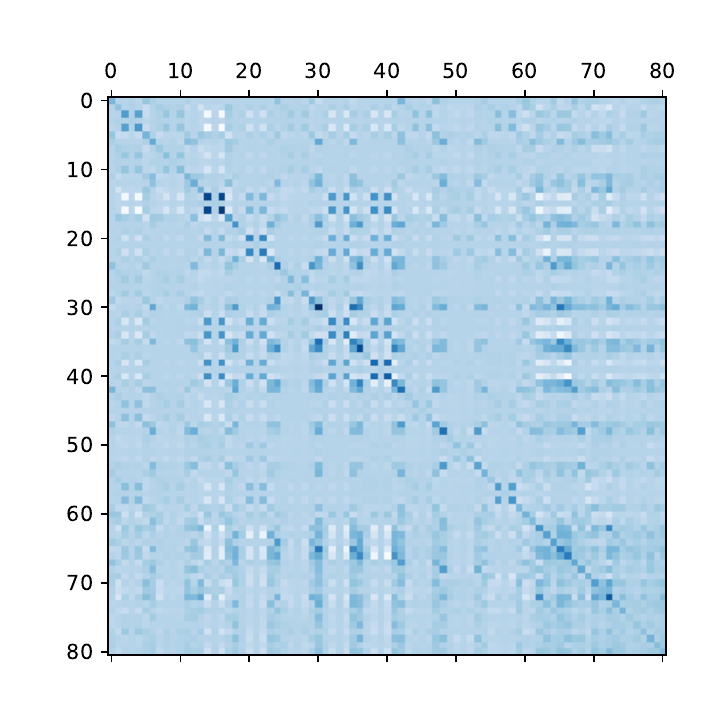}
         \caption{end of epoch 100}
    \end{subfigure}

\centering
\caption{Visualization of snapshots for the \textbf{mean curvature matrix} $M_{\operatorname{SVN}}$, which is the average of the Hessian matrices of all 5 particles at different times during the training process.}
\label{fig:hesscurv}
\end{figure}

\cref{fig:hess1,fig:hess2,fig:hess3,fig:hess4,fig:hess5} illustrate the Hessian matrices for different particles throughout the training process, while \cref{fig:hesscurv} depicts the average curvature matrix used in the anisotropic Gaussian kernel. In all these plots, it can be observed that the Hessian matrix is not particularly expressive at the beginning of the training process. Consequently, we do not expect our SVN method to benefit significantly from more informed gradient steps through curvature information initially. This lack of initial expressiveness means that our method requires a few gradient steps to start effectively utilizing the curvature information and improving its performance. Thus, the SVN method needs some time to settle in before achieving its characteristic convergence speed. While this is a scaled-down example, we are confident that the same principles hold for larger neural networks as well, as seen in \cref{sec: Experiments}.

\end{document}